\documentclass{article}

\usepackage[style=numeric, backend=biber, sorting=none, maxcitenames=2]{biblatex}
\usepackage[english]{babel}
\usepackage{times}
\usepackage[letterpaper,top=2cm,bottom=2cm,left=3cm,right=3cm,marginparwidth=1.75cm]{geometry}
\usepackage{amsmath}
\usepackage{amssymb}
\usepackage{booktabs}
\usepackage{csquotes}
\usepackage{graphicx}
\usepackage{mdframed}
\usepackage{float}
\usepackage{wrapfig}
\usepackage[dvipsnames]{xcolor}
\usepackage[colorlinks=true, allcolors=blue]{hyperref}
\usepackage{lineno}
\usepackage{siunitx}

\def\new#1{#1}
\def\newcite#1{\cite{#1}}

\newcommand{\secref}[1]{Section~\ref{#1}}

\title{Computer Vision for Primate Behavior Analysis in the Wild}
\author{Richard Vogg*$^{1}$, 
Timo Lüddecke*$^{1}$, 
Jonathan Henrich*$^{2}$, 
Sharmita Dey$^{1}$,
Matthias Nuske$^{3}$, \\ 
Valentin Hassler$^{1}$, 
Derek Murphy$^{4,5}$, 
Julia Fischer$^{4,5,6,7}$, 
Julia Ostner$^{6,8,9}$,
Oliver Schülke$^{6,8,9}$, \\ 
Peter M. Kappeler$^{6,10,11}$, 
Claudia Fichtel$^{6,10}$, 
Alexander Gail$^{6,7,12,13}$,
Stefan Treue$^{6,7,12,13}$, \\ 
Hansjörg Scherberger$^{6,7,14,15}$, 
Florentin Wörgötter$^{3,6,7}$,
Alexander S. Ecker$^{1,6,7,16}$}

\date{}

\begin{document}

\maketitle

{\centering\vspace{-12pt}
\small *equal contribution \\
\small $^1$ Institute of Computer Science and Campus Institute Data Science, University of Göttingen \\
\small $^2$ Chairs of Statistics and Econometrics, Faculty of Economics, University of Göttingen \\
\small $^3$ Department for Computational Neuroscience, Third Physics Institute, University of Göttingen\\
\small $^4$ Cognitive Ethology Laboratory, German Primate Center, Leibniz Institute for Primate Research, Göttingen
\\
\small $^{5}$ Department for Primate Cognition, Johann-Friedrich-Blumenbach Institute, University of Göttingen
\\
\small $^6$ Leibniz ScienceCampus, German Primate Center, Leibniz Institute for Primate Research, Göttingen
\\
\small $^7$ Bernstein Center for Computational Neuroscience, University of Göttingen\\
\small $^8$ Behavioral Ecology Department, University of Göttingen\\
\small $^9$ Social Evolution in Primates Group, German Primate Center, Leibniz Institute for Primate Research, Göttingen\\
\small$^{10}$ Behavioral Ecology \& Sociobiology Unit, German Primate Center, Leibniz Institute for Primate Research, Göttingen\\
\small $^{11}$ Department of Sociobiology/Anthropology, University of Göttingen\\
\small $^{12}$ Cognitive Neuroscience Laboratory, German Primate Center, Leibniz Institute for Primate Research, Göttingen\\
\small $^{13}$ Georg-Elias-Müller-Institute of Psychology, University of Göttingen\\
\small $^{14}$ Faculty of Biology and Psychology, University of Göttingen\\
\small $^{15}$ Neurobiology Laboratory, German Primate Center, Leibniz Institute for Primate Research, Göttingen\\
\small $^{16}$ Max Planck Institute for Dynamics and Self-Organization, Göttingen\\
\vspace{12pt}
}

\begin{abstract}
    Advances in computer vision as well as increasingly widespread video-based behavioral monitoring have great potential for transforming how we study animal cognition and behavior. However, there is still a fairly large gap between the exciting prospects and what can actually be achieved in practice today, especially in videos from the wild. With this perspective paper, we want to contribute towards closing this gap, by guiding behavioral scientists in what can be expected from current methods and steering computer vision researchers towards problems that are relevant to advance research in animal behavior. We start with a survey of the state-of-the-art methods for computer vision problems that are directly relevant to the video-based study of animal behavior, including object detection, multi-individual tracking, individual identification, and (inter)action recognition. We then review methods for effort-efficient learning, which is one of the biggest challenges from a practical perspective. Finally, we close with an outlook into the future of the emerging field of computer vision for animal behavior, where we argue that the field should \new{develop approaches to unify detection, tracking, identification and (inter)action recognition in a single, video-based framework}.
\end{abstract}

\begin{refsection}
\section{Introduction}

Monitoring the behavior of animals is an important research task in multiple scientific disciplines ranging from behavioral ecology to neuroscience. 
While behavior is a wide-ranging term, including swarm behavior or movement patterns \cite{couzin2023emerging}, and often studied in laboratories, we want to focus here on individualized actions and interactions recorded in images or videos in free-ranging animals, often in the wild.
Traditionally, the collection of behavioral data has been time-consuming, laborious, or invasive, as it has relied on human observation and manual documentation or equipping animals with biologgers or tracking devices \cite{bateson2021measuring}. In the last decade, the situation has changed as powerful computer vision methods have emerged that have enabled researchers to automate many image recognition tasks. These developments have put fully automated behavioral tracking and analysis of animal behavior on the horizon. Automated annotations using computer vision complement the work of human observers and fundamentally change how data on animal behavior are collected, with important consequences for researchers: Unprecedented amounts of data can be processed and analyzed, something that was infeasible to be done manually; potential biases introduced by researchers who know the research hypothesis could be circumvented; even for smaller quantities of videos researchers can focus on analyzing and interpreting rather than spending weeks or months on manual scoring of videos.

One of the areas where computer vision methods have already been adopted at scale is monitoring animal behavior in laboratory settings \cite{papadakis2012computer, segalin2021mouse, luxem2023open}. Open-source pose estimation frameworks such as DeepLabCut \cite{mathis2018deeplabcut, lauer2022multi}, SLEAP \cite{pereira2022sleap} and others \cite{trex, graving2019deepposekit} have been adopted by a large user community in neuroscience and beyond, and have facilitated an ongoing paradigm shift towards more unrestrained and natural behavior in neuroscience \cite{jordan2024automated}. These tools work remarkably well in a laboratory setting, where the visual complexity of the scene is manageable. However, the situation is more challenging in more natural outdoor environments or in the wild, where complex visual scenes, changing lighting conditions, non-stationary cameras,  occlusion and clutter due to large numbers of animals pose significant challenges to the computer vision machinery \cite{wiltshire2023deepwild}. Moreover, the behaviors that behavioral ecologists are interested in are often complex, not easily deduced from an animal's pose on a single video frame, and often concern interactions between two or more individuals, or between individuals and objects in their environment. The following examples from our own ongoing research work illustrate representative types of problems in which computer vision methods can assist behavioral ecologists: 
First, the identification of individual monkeys from a collection of camera trap images. The co-occurrences of monkeys reveal insights into the social networks of the group of monkeys. 
Second, tracking of multiple monkeys throughout a set of hand-recorded videos and identification of their actions, e.\,g. resting, grooming, sleeping.
Third, the automated analysis of social learning by identifying interactions between individuals who perform certain tasks and observe each other. 

Recently, there have been some successful examples of how to use computer vision techniques to detect, track and identify individual monkeys and their actions in the wild (see \secref{sec:methods}).
While these methods represent an important step towards automated behavior analysis, they usually focus on a single task (e.\,g. detection or tracking) and do not have the same maturity and general usability as models developed for laboratory settings.
The automated recognition of actions and interactions with high accuracy from video recordings in natural environments using computer vision methods is still a challenging and largely unsolved problem. Yet, these settings are what we are particularly interested in.
Making use of currently existing computer vision methods will bring us closer to solving these problems, but we believe it will also require further advances in machine learning and computer vision, \new{unifying several tasks in a single video-based framework.}
Therefore, our aim with this perspective is not only to highlight and discuss current developments in computer vision but also to sketch a path towards future developments.

\begin{figure}[t!]
    \centering
    \vspace{-0.15cm}
    \includegraphics[width=\textwidth]{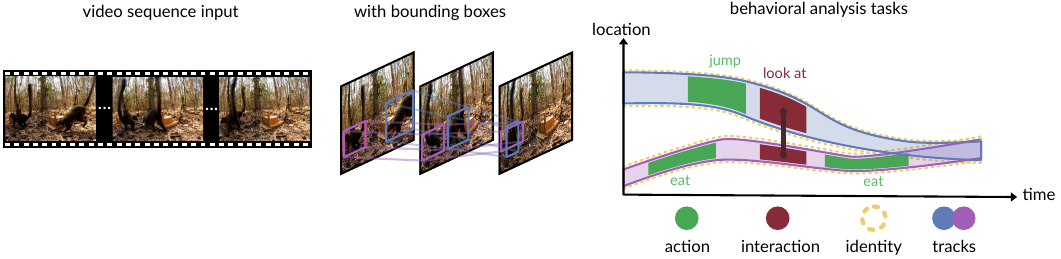}
    \caption{
    Structure of behavior analysis tasks: Starting from a video taken in the wild, the goal is to obtain a structured representation of actions and interactions between the depicted animals while keeping track of their identities. The behavior analysis output on the right shows the behavior of two individuals (blue and violet) over time.}
    \vspace{-0.35cm}
    \label{fig:overview_methods}
\end{figure}

We identify four key tasks on the path towards automated behavior analysis from images and videos. 
These tasks can be combined to obtain a holistic characterization of animal behavior (Fig.~\ref{fig:overview_methods}).
\begin{itemize}
    \item \emph{Animal detection}: To understand the behavior of individual animals, these animals must first be located in images.
    \item \emph{Multi-animal tracking}: For videos, detections of the same individual across multiple frames must be associated with each other to track an individual over time.
    \item \emph{Individual identification}: Many research questions require identifying individuals based on face, body or other features which help to distinguish them.
    \item \emph{Action understanding}: Classifying behaviors based on observable patterns of actions is the core of behavioral analysis. Tasks of interest include the assignment of actions or behaviors to individuals and temporal intervals, as well as the detection of interactions between individuals or between individuals and objects.
    
\end{itemize}

Computer vision researchers have developed a plethora of methods to tackle these tasks. However, these approaches are often tailored to different use cases, mostly involving humans and everyday objects (e.\,g., \cite{dendorfer2020mot20, Lin2014MicrosoftCC, ji2020action, li2020ava}). They do not address the unique requirements of individualized primate behavior analysis in the wild and are therefore not readily usable for primate behavioral analysis. 
Consequently, applied researchers often find themselves without a clear strategy for selecting appropriate methods and establishing systematic protocols for annotation.
To answer the latter, another important question is how to solve the above-mentioned key tasks in the face of limited time and resources. We review current ideas around how to work with completely or partially unlabeled data, labels that are faster to annotate, and how to find the best samples for annotation to improve the model optimally given a certain labeling budget.
Based on our analysis of the state of the art in the key tasks and effort-efficient methods, we highlight promising future directions for computer vision in animal behavior analysis and identify potentially high-impact developments. 

With this paper, we aim to review and structure the field of computer vision for animal behavior researchers with a technical interest. While we write primarily from the perspective of scientists interested in primate behavior in natural environments, many of the discussed principles are also applicable to other taxonomic groups and settings. The paper is directed at both AI researchers and empirical scientists, aiming to make significant contributions in two ways: (1) provide guidance to the deep learning community on critical open problems with practical applications and (2) show biologists and ecologists a path towards transforming their research by AI. 

\new{While there are already several perspective and review papers focusing on animal behavior via 2D or 3D keypoint estimation \newcite{mathis2020deep, pereira2020quantifying, luxem2023open, perez2023cnn}, these applications are almost exclusively in laboratory or fixed background settings. In this perspective, we will focus on methods to overcome challenges when working with videos from the wild. We refer to the abovementioned works for a comprehensive overview of keypoint-based behavioral analysis.}

We review machine learning methods for the key tasks of behavioral analysis (Section~\ref{sec:methods}) and effort-efficient learning (Section~\ref{sec:effort}) in a way that is accessible to researchers from both fields. 
A more detailed technical review of these methods can be found in Appendix~\ref{sec:appCV} and \ref{sec:appEffortEff}, while Appendix~\ref{sec:appFoundations} introduces some computer vision foundations. Our outlook on promising research directions is provided in Section~\ref{sec:future}.

\section{Methods for primate behavior analysis}\label{sec:methods}

In this section, we introduce the computer vision tasks that we expect to be crucial for a holistic analysis of individualized primate behavior: animal detection (2.1), multi-animal tracking (2.2), individual identification (2.3) and action understanding (2.4).
For each of these tasks, we describe the expected inputs and outputs and illustrate the key methods. While some methods operate on static images, others require video input. 
In Appendix~\ref{sec:appCV}, we review all the mentioned methods in more technical detail.

Since the goal is to analyze the behavior of individuals, all tasks introduced in this section require a certain way of representing the presence and position of \emph{object instances} (in our case: individual animals) appearing in an image or video frame. Possible instance representations are bounding boxes, instance segmentation masks and keypoints. 
\emph{Bounding boxes} represent instances by the smallest possible rectangular box that encompasses all visible parts of an animal in an image. Most commonly, axis-aligned bounding boxes are used where the sides of the box are parallel to the image axes. Depending on the actual shape of the animal, axis-aligned bounding boxes can also contain a lot of background but come with the advantage of being fast to annotate. There are variations of this representation, such as amodal bounding boxes, which encompass even the parts of the individual that are not visible (e.\,g. because they are occluded). 
\emph{Instance segmentation masks} indicate for each pixel of the image to which individual it belongs or whether it is part of the background. Labels for instance segmentation can be very time-consuming to acquire, although recent methods show great potential to reduce this time \cite{kirillov2023segment}.
A \emph{keypoint} representation allows a more precise determination of the animal's posture. For each instance, a canonical set of relevant keypoints, such as nose, left hand or right knee is defined. The annotation effort depends on the number of keypoints used, but is usually higher than for bounding boxes.

While many popular toolkits for animal behavior analysis are centered around keypoints \cite{mathis2018deeplabcut, Gler2018DensePoseDH,  pereira2022sleap}, we argue later (see discussion in Section~\ref{sec:future}) that bounding boxes \new{could be} a more suitable intermediate representation for building flexible, generalizable architectures for many applications in the wild.

\subsection{Animal detection}\label{sec:object_detection}
A typical first goal when analyzing animal behavior is to find all animals of interest in an image or video. From a technical perspective, this is an \emph{object detection} problem. Object detection is typically approached at the image level and the most widely used approaches represent individuals as bounding boxes. Training an object detector usually requires images with ground truth labels of bounding boxes. A key challenge for object detection methods is that the number of objects can vary from image to image, as we do not know a priori how many animals there are in an image. Therefore, neural networks, which commonly have a fixed number of parameters, must be equipped with a mechanism to generate a variable number of output bounding boxes. This is typically done by predicting a large number of candidate bounding boxes together with a confidence score, while only bounding boxes with a sufficiently high confidence are kept. 

Object detection methods can be broadly categorized into \emph{single-stage} and \emph{two-stage} methods. In two-stage methods, detection is separated into two steps: Region proposal, where a large number of potential object regions are identified, and classification, where the regions are rated according to their plausibility of containing an object and classified into object categories. This paradigm is employed by the well-known R-CNN family \cite{Girshick2013RichFH, Ren2015FasterRT, Lin2016FeaturePN, He2017MaskR, Dai2016RFCNOD}. Two-stage methods perform very well but are generally fairly complex pipelines. In single-stage methods, bounding box coordinates and confidence scores are predicted without in-between steps. Compared to two-stage methods, single-stage methods are (1) faster (which is important for real-time applications), (2) conceptually simpler, and (3) can be naturally trained in an end-to-end way. Notable examples are Single-Shot Detector (SSD) \cite{liu2016ssd}, the YOLO-series \cite{Redmon2016YOLO9000BF, Redmon2018YOLOv3AI, Ge2021YOLOXEY} and heatmap-based detectors \cite{Law2018CornerNetDO, Duan2019CenterNetKT, Zhou2019ObjectsAP, tian2019fcos}, where bounding boxes are predicted with the help of heatmaps for pre-defined object parts, such as corners or centers. The methods described so far are all based on Convolutional Neural Networks. More recently, single-stage object detection methods based on the transformer network architecture \cite{Vaswani2017AttentionIA} have gained popularity. In this paradigm, objects are represented by so-called object queries, i.e. numerical vectors, that extract information from the input image through a transformer network. These object queries are then transformed into bounding boxes with corresponding confidence scores which represent detected objects. Notable examples following this paradigm include the detection transformer (DETR) \cite{Carion2020EndtoEndOD} and its follow-up works \cite{Zhu2020DeformableDD, Li2022DNDETRAD, Liu2022DABDETRDA, Zhang2022DINODW}. 

\new{
In the recent years, detection methods have been adapted for non-human primates in the wild. There are successful applications of both two-stage detectors \newcite{ schneider2018deep, reid_face} as well as YOLO and DETR-based one-stage detectors \newcite{yang2019great, roy2023wildect, yang2023dynamic}.
Moreover, several datasets provide bounding box annotations for primates in the wild.
The BaboonLand dataset \newcite{duporge2024baboonland} includes \num[group-digits=integer]{30000} labeled images of baboons captured with drones for multi-animal tracking. The CCR dataset also provides multi-animal tracks in the wild \newcite{Bain2019CountCA}, which result in almost \num[group-digits=integer]{1000000} labeled images. The PanAf20K \newcite{brookes2024panaf20k} and ChimpAct \newcite{ma2023chimpact} action understanding datasets each provide more than \num[group-digits=integer]{15000} labeled images of chimpanzees and gorillas captured in the wild and in the zoo respectively. 
Further, there are numerous datasets for pose estimation \newcite{ye2022superanimal, yao2023openmonkeychallenge, Desai2022OpenApePoseAD, Labuguen2020MacaquePoseAN}.}
\new{We review the object detection methods in more technical detail, as well as additional datasets for object detection and pose estimation in Appendix \ref{app:detect}}.

\subsection{Multi-animal tracking}
\label{sec:tracking}
Object detection methods can localize animals of interest in images. However, researchers frequently record video footage of animal behavior. In this case, they need a model that will not only find all animals in a given video frame but also keep track of them over time. This tracking is crucial for observing long-term behavioral patterns and the assignment of instances and entire sequences of actions to the same animal over time. 

The input of a tracking model is always a video, typically broken down into single frames in a temporal sequence. The output is a list of detections (most commonly bounding boxes) and track IDs that link the detected instances across video frames. A good tracking algorithm will be able to follow an individual through the sequence and assign the same identity to each detection of that individual. When the detections are represented as bounding boxes, this problem is referred to as \emph{multi-object tracking}. Although tracking algorithms based on other instance representations exist (e.\,g. Video Instance Segmentation or keypoint-based tracking), they often solve multi-object tracking as a first step or have to solve it implicitly. For this reason and because multi-object tracking is the most widely studied and best-understood paradigm, we will focus on it here.

The tracking procedure has so far usually been partitioned into two phases \cite{ciaparrone_deep_2020}: detection and association. In the detection phase, the primary aim is to identify and locate objects of interest within frames, as described above. The association phase links the detected objects across frames to produce trajectories over time for each object. This is accomplished through the detected locations and the extraction of descriptive features, such as motion and appearance, from the detected objects. The motion features estimate the position of an object in future frames, while the appearance features represent its visual characteristics. After feature extraction, an affinity cost is calculated. The affinity cost measures the pairwise similarity between the detected objects across frames based on their location, motion and appearance. It is then used to link the detected objects across successive frames to generate object trajectories over time. Existing multi-object tracking methods can be categorized into two groups: (1) The classical \emph{tracking-by-detection} approach described above, where the detection and association phase are independent of each other, and (2) more recent \emph{tracking-by-query} methods with interdependent detection and association phases. 

A prime example of the classical tracking-by-detection framework is the Simple Online Real-time Tracker (SORT) model \cite{bewley2016simple}. In the SORT model, the detections of each frame are collected independently of each other. The spatial shift between detections of successive frames is then used to extract motion features, which are used to calculate the expected location of objects on the next frame. Detections are then linked across frames based on the overlap between the expected location of objects and the actual detections. Follow-up works improved the SORT model, for example by additionally including appearance features \cite{zhang2022bytetrack,cao2022observation,maggiolino2023deep} or employing more elaborate association strategies \cite{zhang2022bytetrack,aharon2022bot}.

More recently, frameworks have emerged that follow the tracking-by-query paradigm where the detection and association phases are interlinked, rather than independent. In this setup, the results of the association phase, such as motion or appearance features, help improve the results of the detection phase, and vice versa. The most prominent example of this paradigm is the multiple-object tracking with transformer (MOTR) model \cite{zeng2022motr} and its follow-up works \cite{Zhang2022MOTRv2BE, yu2023motrv3, yan2023bridging}. MOTR uses the DETR detection framework described above and introduces the concept of track queries in addition to object queries. While object queries are used to detect newly emerging objects, as is done in DETR, track queries represent tracked objects in terms of their motion and appearance features and are used to produce tracks of objects. Object and track queries interact with each other through the transformer design which enables information exchange between detection and association.

\new{Several datasets have been introduced that provide multi-animal tracking labels in the context of primates. For example, the CCR dataset \newcite{Bain2019CountCA} provides roughly ten hours of chimpanzee multi-animal tracks in the wild. The recently published BaboonLand dataset \newcite{duporge2024baboonland} contains multiple videos with multi-animal tracking annotations, spanning half an hour. PanAf20K \newcite{brookes2024panaf20k} and ChimpAct \newcite{ma2023chimpact} each contain roughly two hours of labeled multi-animal tracks from 500 and 163 videos, respectively. Apart from these dataset contributions, there are also some recent works presenting methods for multi-animal tracking in the wild. For example, \textcite{pineda2023deep} employ a standard tracking-by-detection framework for chimpanzees and bonobos in the wild. \textcite{wiltshire2023deepwild} provide evidence that DeepLabCut \cite{mathis2018deeplabcut} can be used to track chimpanzees and bonobos in their natural habitats. However, they also acknowledge significant challenges due to the variability of videos in the wild, particularly with novel test videos.}

\subsection{Individual identification}
Individual identification is an essential step for many studies on animal behavior and ecology, for example in studies of sociality \cite{cam_trap-temp, schofield2023automated} or for population estimation \cite{cam_trap-pop}.
It describes the task of distinguishing between individuals in an animal group by consistently assigning unique identity labels across different images or videos. The label is typically a pre-assigned identity, e.\,g. a name or ID code. Individual identification is different from multi-object tracking, where individuals are assigned a new label for each video/scene in which they appear. 

There are two main approaches to individual identification: \emph{closed set} and \emph{open set} identification \cite{vidal2021perspectives}. In closed set identification, all individuals are known beforehand and need only be recognized again in new images, while in open set identification previously unseen individuals (e.\,g. from unhabituated groups) can appear in the data and must be distinguished from known individuals. The simplest way to tackle closed set identification is to train an object detection or classification method to directly output the identity of an animal. \new{This approach has been extensively employed in primate studies, where networks have been applied to images of the entire body \cite{marks2022deep}, the face \cite{reid_face, reid_face3, freytag2016chimpanzee, brust2017towards, brookes2020dataset, tieo2023mandrillus}, or a larger region of interest containing multiple animals \cite{Bain2019CountCA}. Recently, there has been evidence that deep metric learning for the identification of animals outperforms classification approaches \cite{vidal2021perspectives}. In this paradigm, the goal is to learn features that can be used to distinguish individuals from each other. Many identification methods that rely on deep metric learning were originally introduced in the context of human face recognition \cite{parkhi2015deep, schroff2015facenet, liu2017sphereface, deng2019arcface} and subsequently adapted to primate settings \cite{reid_face2, charpentier2020same}}. One advantage of deep metric learning is that unknown individuals can be identified based on feature distance. More precisely, when the calculated feature of a newly detected individual is highly dissimilar to the features associated with known individuals, this individual can be classified as unknown. Therefore, deep metric learning is a promising avenue to tackle the challenging task of open set individual identification.

\new{The identification of individuals is a prerequisite for tackling many social behavior questions and multiple labeled datasets have been published for various primate species. Among these datasets are ChimpAct \newcite{ma2023chimpact} with \num[group-digits=integer]{160000} frames and 23 individuals, the Mandrillus Face Database \newcite{tieo2023mandrillus} with \num[group-digits=integer]{29495} face images of 397 mandrills, studies on chimpanzees \newcite{schofield2019chimpanzee} with 10 million face detections of 23 individuals, gorillas \newcite{brookes2022evaluating} with \num[group-digits=integer]{5428} face detections of seven individuals, and Japanese macaques \newcite{paulet2024deep} with \num[group-digits=integer]{2485} frames of the faces of 42 individuals. Furthermore, \textcite{reid_face} introduced a dataset containing \num[group-digits=integer]{11637} images of 280 individuals from 14 different primate species. In practice, models trained on these datasets could serve as a starting point to obtain models for one's own identification setting, which might reduce the amount of labeled data required (see transfer learning in \secref{sec:effort}). Transferring information from other identification datasets has been demonstrated for human identification tasks \newcite{peng2016unsupervised, li2018adaptation}. Furthermore, these datasets can serve as benchmarks to develop and systematically compare new identification methods.}

\subsection{Action understanding}\label{sec:action_understanding}
\label{sec:action}

The animal localization methods discussed in the previous sections (detection and tracking) are crucial prerequisites. Now we focus on methods for action understanding. \new{In the animal behavior literature, action understanding is mostly treated as a secondary task that is approached by first extracting keypoint tracks of all animals in the scene, which are then used to determine the animals' actions (reviewed, e.\,g. by \textcite{luxem2023open, mathis2020primer, pereira2020quantifying, couzin2023emerging}). In contrast, human action understanding is also an active research area in computer vision, where methods that operate directly on video snippets dominate. As human action understanding faces many of the same challenges as observing animals in the wild, this line of research warrants particular attention.} 

\begin{figure}[t!]
    \centering
    \includegraphics[width=11cm]{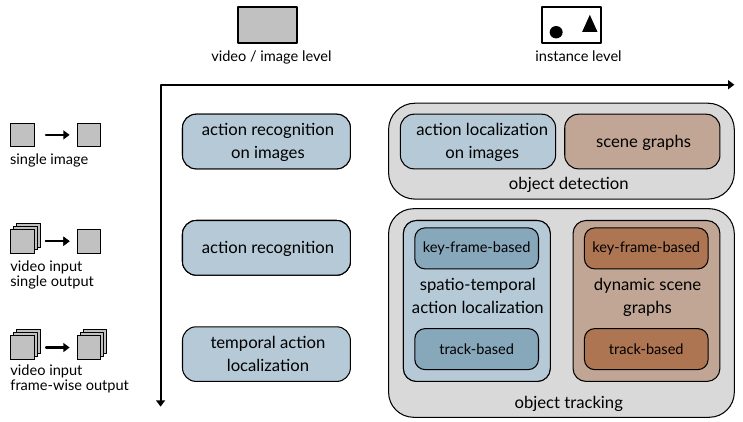}
    \caption{Categories of action and interaction recognition tasks based on input and output temporal format as well as spatial granularity.}
    \label{fig:actions}
\end{figure}

These methods for action understanding differ mainly in whether and how they localize actions in space and time (illustrated in Fig. \ref{fig:actions}). For video inputs, they can be divided into four main categories: (1) \emph{action recognition} (also known as action classification), (2) \emph{temporal action detection} (also known as temporal action localization), (3) \emph{spatio-temporal action detection} (also known as spatio-temporal action localization) and (4) \emph{dynamic scene graph generation}. To exemplify what these four methods do, suppose that there is a video of a monkey feeding from a bowl. An \emph{action recognition} method would assign the label ``feeding'' to the entire video. A more precise output can be obtained using a \emph{temporal action detection} method which would identify the temporal interval during which the monkey is feeding. Both of these methods generate video-level output that does not refer to individual animals. These methods assume that the input videos focus on a single individual, i.\,e. individuals have already been localized. In contrast, a \emph{spatio-temporal action detection} method would localize the monkey using bounding boxes on individual video frames (frame-based approach) or a track (track-based approach). The action ``feeding'' would then be assigned to the respective bounding boxes (frame-based approach) or sub-tracks (track-based approach) where the monkey feeds. Similarly, a \emph{dynamic scene graph generation} method would localize the monkey and the bowl by means of bounding boxes or tracks. The action ``feeding'' would then be assigned to the respective bounding boxes or sub-tracks in the form of a relationship triplet (monkey--feeding from--bowl). It should be noted that frame-based approaches still leverage temporal information for the prediction. However, predictions are made separately for each frame, which is why it is not possible to directly infer whether two actions on different frames belonging to the same animal represent the same action instance or whether the action has been interrupted in between (which is relevant if one is interested in counting how often a particular action was performed). To achieve this, a tracking method would have to be applied additionally. Figure \ref{fig:actions} illustrates the action understanding methods with regard to their input and output type.

Advances in action understanding methods have been driven largely by the development of more and more powerful neural networks (``backbones'') for video processing. While early approaches relied primarily on 3D convolutions \cite{qiu2017learning, lee2021diverse, feichtenhofer2019slowfast}, there has been a shift towards transformer networks \cite{arnab2021vivit, bertasius2021space, yan2022multiview, piergiovanni2023rethinking, fan2021multiscale, ryali2023hiera, wu2022memvit, li2022uniformer, li2022uniformerv2}, enabling the exchange of information across spatially and temporally distant regions of the input video. State-of-the-art action understanding results can often be achieved by simply employing a powerful video backbone with a classifier network on the entire video clip (action recognition) or a spatially cropped version (spatio-temporal action detection). The region for spatial cropping is usually determined by employing a separate detector for actors. Explicitly modeling motion \cite{patrick2021keeping, long2022stand} or action-related components \cite{materzynska2020something, tian2022ean, faure2023holistic}, such as actors and objects involved, for action recognition and spatio-temporal action detection has also been attempted, but often performs worse than simply using strong general video processing backbone without any bells and whistles.
Temporal action detection methods are conceptually similar to object detection methods, with the difference being that temporal instead of spatial boundaries are predicted. Just as in object detection, there is a trend towards single-stage methods that process the input using some form of video backbone and directly predict action boundaries and confidence scores without any proposal generation step in between \cite{cheng2022tallformer, zhang2022actionformer, tang2023temporalmaxer, shi2023tridet}. 

The methods described so far assign actions either to entire videos, subsequences of videos or detected individuals. But researchers are often interested in \emph{interactions} between two individuals or individuals and specific objects: for instance, who is grooming or looking at whom? Which individual is manipulating which object? Such interactions can be described by a \emph{scene graph}, where edges express interactions between entities (objects/individuals). Generating dynamic scene graphs is an emerging field of research. Existing methods \cite{wang2023cross, feng2023exploiting, gao2022classification, wei2023defense} typically involve the following steps: First, features are extracted for entities (e.\,g. monkey and bowl) and relations (e.\,g. feeding). The entity features are usually obtained from image or video backbones, while relations are, for example, represented by the relative motion of the involved entities. In a second step, the extracted features are further processed using some form of spatio-temporal reasoning. Finally, the features are used to separately predict the entity and relation classes.

\new{
A number of groups have recently started tackling automatic action understanding for primates, both in terms of adapting methods from computer vision as well as collecting, annotating and publishing datasets. Some studies perform behavioral analysis using image-based backbones, for example for basic movement actions \newcite{sakib2020visual} or postural recognition \newcite{lei2022postural}. Due to the recent emergence of several video-based action understanding benchmarks, there has been a trend towards video-based methods \newcite{bain2021automated, brookes2023triple, brookes2024chimpvlm, zhao2024videoprism}. For example, \textcite{bain2021automated} employ 3D-convolutions on bounding boxes of individuals over time for spatio-temporally localized behavior recognition. Video-based action understanding datasets that contain primates include the following: MammalNet \newcite{chen2023mammalnet} (\num[group-digits=integer]{18000} videos, 539~h) comes with action recognition annotations for several mammal species. ChimpBehave \newcite{fuchs2024forest} (\num[group-digits=integer]{1121} videos, 2~h) consists of videos containing a single chimpanzee annotated with spatio-temporal action detection labels. PanAf20K \newcite{brookes2024panaf20k} (\num[group-digits=integer]{500} videos, 2~h) and ChimpAct \newcite{ma2023chimpact} (\num[group-digits=integer]{163} videos, 2~h) both include spatio-temporal action detection annotations of groups of monkeys. PanAf20K additionally provides multi-label action recognition annotations for a larger set of videos (\num[group-digits=integer]{20000} videos, 80~h). }
For more technical details and an in-depth discussion of action and interaction understanding, we refer to Appendix \ref{app:action}.

\section{Methods for effort-efficient learning}
\label{sec:effort}

\begin{figure}
    \centering
    \includegraphics[width=12cm]{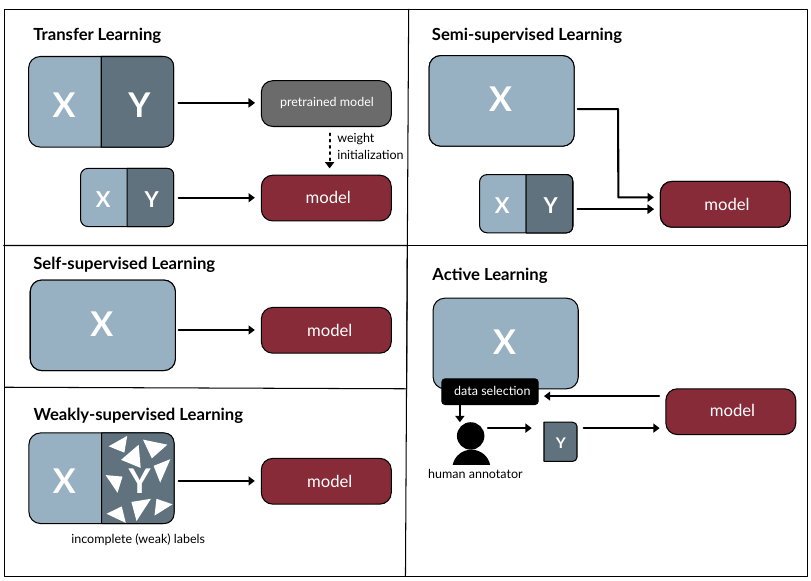}
    \caption{Schematic overview of effort-efficient learning techniques.}
    \label{fig:effort_efficient}
\end{figure}

The methods presented in the previous section have been developed using (usually) large, diverse image or video datasets with high-quality annotations. Thus, while there has been a lot of progress and many strong methods exist in principle, they cannot be easily applied to situations where no or only a few labels exist -- as is the case in animal behavior. Annotating images or videos with bounding boxes, keypoints, segmentation masks or interactions takes a considerable amount of time and often labeling budgets are limited. 
We therefore now turn to effort efficiency: Researchers would like to maximize model performance while minimizing manual annotation costs. Which techniques can we use to make the most of available labels or even of unlabeled data? Figure \ref{fig:effort_efficient} provides a schematic overview of some of the most promising effort-efficient learning techniques. Note that the boundaries between these categories are fluent and some methods do not strictly fall into only a single category, but are hybrid approaches. It helps, however, to look at the approaches in isolation as they deal with different types of input data and model architectures. Below we provide a high-level overview of these methods. 
In Appendix~\ref{sec:appEffortEff} we review the currently used methodologies for effort-efficient learning in more technical detail.

The most traditional effort-efficient learning method is \emph{transfer learning}. The idea is to transfer knowledge gained from one task to enhance the performance on a related task. Traditionally, a model trained on an annotated dataset with a wide range of objects, such as ImageNet \cite{Russakovsky2014ImageNetLS} or MS-COCO \cite{Lin2014MicrosoftCC}, is fine-tuned with data from the task of interest. Since knowledge contained in the original model can be leveraged, this allows for training good models even when only a small number of labels is available. \new{The effectivity of transfer learning depends on various factors, including the type of tasks being used \cite{zamir2018taskonomy,mensink2021factors}, diversity of the data \cite{nayman2024diverse,mensink2021factors} and robustness of the features \cite{salman2020adversarially}.}
\new{Within the context of animal behavior, SuperAnimal \cite{ye2022superanimal} trains a common pose estimation model across a large set of animal species and has been shown to be data-efficient when fine-tuning to new pose estimation datsets.}

\emph{Self-supervised learning} models learn meaningful representations from large, unlabeled datasets. One of the most popular approaches is masked autoencoders \cite{he2022masked}, where parts of the image are masked out and the model has to learn to restore the image. Another line of self-supervised models is contrastive learning \cite{chen2020simple, chen2021exploring}, where two similar versions of an input image are created (e.\,g. by random cropping, flipping, color changes) and the model has to learn that they show the same image content while differentiating it from other images. Both ideas require the model to learn a general understanding of what an image shows. After self-supervised pre-training on a large unlabeled dataset, the models are adapted to the target task using transfer learning with a small labeled dataset. This can yield better results than supervised pre-training \cite{he2022masked}.

\emph{Weakly supervised learning} is used in scenarios where perfect annotations are not available. Models are trained using incomplete, inaccurate or cheaper labels. Especially the latter is relevant for our animal behavior models, as we can train a model without having to go through the full annotation effort. Annotations of image-level labels, bounding boxes, keypoints and instance segmentations are increasingly more fine-grained and expensive. For example, for individual identification, we need bounding box labels, but drawing bounding boxes is more time-consuming than creating more coarse-grained image-level labels (e.\,g. simply stating which individuals are present in the image). Weakly supervised methods aim to find ways to train a fine-grained model with coarse-grained labels, thus minimizing annotation time.
For great ape detection, \textcite{yang2023dynamic} employ weakly-supervised learning based on pseudo labeling and curriculum learning and demonstrate a superior performance compared to supervised baselines.

\emph{Semi-supervised learning} is a strategy that utilizes both labeled and unlabeled data for model training. A basic strategy is to train a first model using the labeled data and use this to create pseudo labels on the unlabeled data, with which a better model can be trained.
In the popular architectures, MixMatch \cite{berthelot2019mixmatch} and FixMatch \cite{sohn2020fixmatch}, the pseudo labeling strategy is combined with consistency-based methods, where the model learns to consistently recognize an image as the same, even if it is flipped, cropped, has color changes or other augmentations applied to it.
\new{It has to be noted that the out-of-domain generalization capabilities of semi-supervised learning models have been questioned, especially when the unlabelled part of the data comes from a different domain \newcite{oliver2018realistic}.}

\new{\emph{Data-centric learning} methods have become a hot topic in machine learning research that is also highly relevant for understanding animal behavior. Data-centric methods shift the focus from designing models to selecting high-quality data samples that are most beneficial for training. This can substantially improve the performance as (semi-)automatically collected datasets are often biased and contain redundant, low-quality or falsely labelled samples \cite{zha2023data}.}
\new{Recent papers showed that a smart selection of training data can improve the performance of vision-language models for generic image recognition tasks \cite{gadre2024datacomp,xu2023metaclip,fang2023datafiltering}. Such strategies could also be employed specifically for primate-related tasks.}
For instance, when annotating behavioral videos, randomly sampling frames could result in many frames without relevant individuals and actions, repeated samples of the same individual, too simple or too complicated frames. 
\new{Active learning is a data-centric technique where a model decides which samples are most beneficial to be annotated. These selected samples are then annotated by the user (or an oracle) and the model is re-trained on the extended dataset which contains these new samples. This process can be repeated multiple times.}
While active learning methods mostly focused on image classification tasks, there exist first applications in animal behavior analysis \cite{gunel2019deepfly3d, tillmann2024soid, li2022openlabcluster}.
Two main strategies are used: Uncertainty-based methods look for samples where the current model shows low confidence \cite{yoo2019learning}; diversity-based methods select samples for annotation such that the entire dataset is well represented \cite{sener2017active}.
\new{In active learning, repeated evaluation on the same validation set can lead to an underestimation of the generalization error and a false impression of robustness. Hence, it is important update the validation set with sufficiently new samples and (this holds for all machine learning methods) to use sufficiently distinct samples for computing the test error.}

\emph{Synthetic training data} is another approach to avoiding manual annotation. This usually involves data produced using 3D renderers, where labeled samples can be generated at no cost once a simulation model has been created (e.\,g. Kubric \cite{greff2022kubric}). Depending on the quality of the simulated data, it might be necessary to explicitly adapt to real-world data, for example using transfer learning. In some cases, directly using the model trained on synthetic images on natural images has proven to be successful. This involves optical flow \cite{Dosovitskiy2015FlowNetLO} and point tracking \cite{karaev2023cotracker, doersch2023tapir}. \new{Recently, Plum et~al. \cite{plum2023replicant} generated synthetic training data for ant behavior analysis.}

\emph{Cross-modal learning} has recently emerged as a popular paradigm where observations from a second modality serve as supervisory signals. For instance, the vision-language model CLIP \cite{Radford2021LearningTV} uses the text accompanying images on websites as ``labels'' for the images. 
\new{It has been shown that the CLIP encoder provides a useful basis for a diverse set of computer vision tasks, either through fine-tuning or even as a frozen encoder (i.e. without changing weights). The set of tasks involves general \newcite{ju2022prompting,lin2022frozen} and animal \newcite{jing2023category} video action recognition, keypoint detection \cite{yang2023unipose}, semantic image segmentation \newcite{lueddecke22_cvpr}, point cloud recognition \newcite{zhang2022pointclip} and camera-trap image retrieval \newcite{gabeff2024wildclip}}.
\new{Beyond that}, researchers could collect audio recordings, thermal images, accelerometer data, GPS information, weather data and more, and use the data from these extra modalities as labels for video data in a cross-modal learning setting. Incorporating audio information into a machine vision model can lead to good performance on tasks relevant to researchers interested in animal behavior, as demonstrated by Bain and colleagues \cite{bain2021automated} who used such an approach to identify actions performed by wild chimpanzees.

Finally, annotation efficiency can be improved by using collaborative efforts between humans and already existing models. Integration of deep learning models into annotation workflows and the design of user-friendly tools are means to enhance the efficiency of collaborative annotation processes.

\section{Avenues for Future Research}
\label{sec:future}

\begin{figure}
    \centering
    \includegraphics[width=\textwidth]{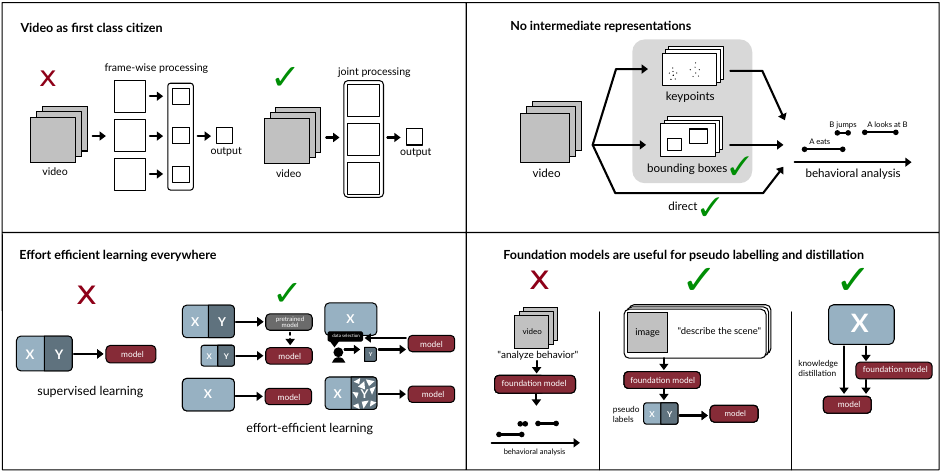}
    \caption{Selection of important developments in computer vision and their impact on behavioral analysis.}
    \label{fig:future}
\end{figure}

\new{In the previous sections we have reviewed what we consider the fundamentals for developing a path from primate videos to automated behavioral analysis in the wild. This development is still in an early stage, and many problems and questions remain unsolved: How much data needs to be annotated for which task? How can we make models and approaches usable across species and tasks? Which solutions can we develop particularly for videos from the wild, dealing with occlusion, variable lighting conditions and backgrounds, as well as unrestricted movements of the primates? How do we design simple yet flexible and extensible frameworks that encompass all relevant tasks from detection and tracking to individual identification and (inter)action recognition without complex multi-stage architectures?
While researchers in computer vision and behavioral sciences} developed approaches to tackle these problems, not all of them have received the same degree of attention. For instance, \new{in the context of detection and tracking, the computer vision community has focused much more on image-based approaches than on true video processing}, mostly for reasons of computational complexity and memory. \new{In addition, existing video-based methods often do not provide the spatio-temporal output required for fine-grained individualized behavioral analysis}. Research in computer vision is driven strongly by the availability of benchmarks and (annotated) datasets. The focus has often been on developing architectures that work well in certain domains (e.\,g. pedestrian tracking) given the constraints of the size and types of datasets available. Comparably less effort has gone into researching how to obtain the necessary labels in the most efficient way in a multi-task setting, how to reuse labels across tasks and domains, and how to design architectures that facilitate such reuse. In this final section, we want to highlight what we consider to be the most important open questions in computer vision for animal behavior (Fig.~\ref{fig:future}).

\paragraph{Video as a first-class citizen.}

Actions and interactions -- the primary quantity of interest in behavioral science -- are inherently spatio-temporal. We argue that, for this reason, video should be treated as a first-class citizen. 
\new{In the last decade, computer vision research arguably put more effort into image-based tasks such as image classification, object detection or image segmentation than into video-based tasks.}
As a result, even inherently video-based problems are often addressed by first performing frame-level image recognition and then merging/tracking results over time. A prime example of this phenomenon is the tracking-by-detection approach (see~\secref{sec:tracking}) with independent detection and association phases: image-based object detectors have been optimized for years and large, high-quality annotated datasets exist, but datasets with densely annotated object tracks in video are rare and comparably small. As a result, tracking approaches, which are based on state-of-the-art detectors, outperform methods directly trained on object tracks on typical pedestrian-tracking datasets. However, these image-based approaches likely lead to locally, rather than globally optimal solutions, especially when it comes to classifying animal behavior. That is, frame-based recognition cannot deal as well with low-level image corruptions like motion blur or poor lighting conditions, has fundamental difficulties with reasoning about appearance changes caused by non-rigid body movements and is more error-prone when it comes to clutter or occlusion of similarly looking individuals. All of these issues are in principle also present in tasks involving tracking humans moving through a space, but are far more pronounced when using videos of animals, because animals look more similar, move faster and more non-linearly. The community has recently started to recognize and tackle these challenges. For instance, the DanceTrack benchmark \cite{sun2022dancetrack} brought into focus the challenges of dealing with similar-looking individuals, complex motion and occlusion. From our perspective, a significant proportion of these difficulties could be alleviated by shifting from image-level to video-level methods. Recent query-based models where detections are obtained by also taking into account past information of already-tracked objects are a step in the right direction \cite{zeng2022motr, Zhang2022MOTRv2BE, yu2023motrv3}. However, these approaches still process the video in a frame-by-frame manner, thus preventing the learning of video-level features. 

\new{In contrast to tracking, action understanding tasks have already been tackled with video-level processing by the computer vision community (see~\secref{sec:action_understanding}).
For example, a lot of research is done on action recognition and temporal action detection. 
But these methods are not readily usable by behavioral scientists for fine-grained individualized behavioral analysis: One would first have to detect and track individuals, extract spatiotemporal crops around each individual from videos for classification, and then temporally merge those for actions longer than the typical temporal horizon of action recognition methods.
Similarly, methods that do provide spatially localized output often only generate results independently for individual frames. To obtain results for an entire video, a separate tracking method needs to be integrated, which results in complicated pipelines.} 
When studying the behavior of a group of primates, researchers want to localize actions and interactions in space and time, and identify which individuals are involved. Formally speaking, we want to infer a scene graph, where nodes represent individuals (in space and time) and edges represent their actions and interactions. Thus, we ultimately need a framework that consists of a strong spatio-temporal backbone, ideally trained self- or semi-supervised with modular heads that output tracks for individuals, identifications, pixel-level segmentations and/or keypoints, spatio-temporally localized actions and interactions, or subsets/combinations of those. Such unified models increase the practical usability and ensure that systems can be easily further developed, avoiding the pitfalls of overly complicated pipelines that often prove to be dead ends.
\new{Recently, some groups proposed approaches that move in this direction, such as joint tracking and action detection \cite{zhao2022tuber, gritsenko2023end} or spatio-temporal scene graphs \cite{gao2022classification, wei2023defense}. Although these methods are still not nearly as mature as their thoroughly studied single-task counterparts, it shows that part of the community acknowledges the need to make this transition and is moving in this direction. Moreover, the development of challenging, comprehensive behavioral benchmarks such as the recently published ChimpAct \cite{ma2023chimpact} and PanAf20K \cite{brookes2024panaf20k} can help researchers take a more holistic perspective on action understanding. 
Although the compute and memory requirements of video-based methods are certainly higher than for image-based methods, we expect that current trends towards memory-efficient architectures and rapid progress in hardware development will quickly pave the way for widespread usage of video-based methods. Even today, state-of-the-art video backbones \cite{fan2021multiscale, ryali2023hiera, wang2023masked} can be trained with reasonably large batch sizes on a single GPU (more details at the end of Appendix~\ref{app:action}). Given the unexplored potential of video-based methods and their proven superiority over image-based processing, we believe that a shift in focus is highly warranted.}

\paragraph{Intermediate representations.}

Pose estimation has become a popular first step and intermediate representation in video-based analysis of animal behavior -- in particular in lab-based settings, e.\,g. in neuroscience. Pose estimation as a first step has been facilitated by impressive open-source frameworks such as DeepLabCut \cite{mathis2018deeplabcut}, SLEAP \cite{pereira2022sleap} and DeepPoseKit \cite{graving2019deepposekit}, which provide excellent tools for data annotation, are well-documented, relatively easy to use for non-experts and work remarkably well in lab settings or when overfit to individual video sequences. Other works built upon these frameworks by providing behavioral analysis tools for laboratory settings that are based on features derived from tracked keypoints \cite{nilsson2020simple, segalin2021mouse, sun2023mabe22}.

Keypoints clearly contain useful information for some downstream tasks such as understanding kinematics or certain actions which involve the movement of body parts, e.\,g. fingers. 
However, they do not lend themselves as naturally to other important problems such as individual identification, or classifying certain actions. For example, the action ``monkey feeding'' can be more unambiguously detected using visual information (e.\,g. food in the hand). Thus, while being helpful sometimes, the extraction of visual information from an image or video \new{often} does not necessitate keypoints.
\new{While all intermediate representations have their advantages and disadvantages, we argue that there may be a few} reasons for preferring bounding boxes as the intermediate representations around which to build generalizable architectures: First, bounding boxes are faster and easier to obtain compared with other instance representations.\footnote{\new{In MS-COCO \cite{lin2014microsoft}, a polygon for a single object has on average 24 points, humans are defined by 17 keypoints. Bounding boxes contain only two points. Of course, depending on the task, a smaller number of keypoints might be sufficient, lowering the annotation effort.}} Second, as deep learning models become more and more powerful, we expect them to be able to extract the relevant features for tasks such as action understanding directly from the input image or video, without the need for complex instance representations. If more fine-grained instance representations are needed, they can be easily added as a second step when the instances are already detected and localized. Third, state-of-the-art action understanding methods already rely predominantly on bounding boxes as instance representations. This observation mirrors a general trend: Over the history of deep learning, models using intermediate representations have disappeared and were replaced by models that are trained ``end-to-end'' directly on the task of interest (e.\,g. action understanding). Until 2012, image recognition classifiers were trained on manually-defined visual representations (bag-of-visual words \cite{Csurka2002VisualCW}, histogram of oriented gradients \cite{Dalal2005HistogramsOO}). The seminal AlexNet \cite{krizhevsky2012imagenet} showed that end-to-end training directly on the image classification task outperforms handcrafted intermediate representations. 
\new{One noteworthy attempt of studying animal behavior in the lab without intermediate representations is DeepEthogram \cite{bohnslav2021deepethogram}.}
\new{Arguably, there are challenges of bounding boxes such as a bad coverage for irregular shaped objects (e.\,g. an elongated object over the diagonal) or referencing underlying object properly in case of occlusion \cite{zhou2023rethinking}.}
\new{Despite these shortcomings, we} expect this transition to also happen in animal behavior analysis, which currently relies heavily on some form of intermediate representation. A key question will be if labeled data is necessary as a driver or if features can also be learned in an unsupervised way -- our next topic.

\paragraph{Effort-efficient learning everywhere.}

Progress in image recognition was initially driven by end-to-end training on large collections of labeled data such as ImageNet \cite{Russakovsky2014ImageNetLS} and MS-COCO \cite{Lin2014MicrosoftCC}, but the image characteristics of such commonly used datasets differ quite a bit from those encountered when studying animal behavior in the wild, limiting the power of transfer learning from such datasets \new{\cite{mensink2021factors}}. Collecting and annotating large-scale ``in-domain'' datasets (i.\,e. with relevant species in the appropriate environments and contexts) is costly. Hence, such datasets might not be available any time soon for many relevant animal behavior tasks. However, recent years have brought a lot of progress in training models without the need for manual data annotation, as discussed in \secref{sec:effort}. In our view, a particularly promising research direction in the context of animal behavior is self-supervised learning in combination with transfer learning.

Self-supervised learning is promising because large collections of video footage of animal behavior are increasingly available as researchers employ video recordings for behavioral observations or set up camera traps in the wild. Such datasets could be used to pre-train a backbone in-domain. However, it remains to be investigated which pre-training objectives yield features that are useful for the downstream task: it is not at all clear whether self-supervised learning approaches developed on ImageNet or YouTube videos will generalize to the problems posed by analyzing animal behavior. For instance, the self-supervised paradigms developed using data from ImageNet do not readily generalize to other image domains, because ImageNet images are mostly focused on a single salient object and work by applying label-preserving transformations to these images. Such approaches may not easily generalize to settings where there are many similar-looking individuals close together. Current video-based self-supervised learning strategies are largely inspired by their counterparts in the image domain. The most successful models rely on masked feature modeling \cite{wang2023masked, tong2022videomae, wang2023videomae} and video-language contrastive learning \cite{wang2022internvideo}. Putting animal behavior recognition on the agenda for evaluating self-supervised video representation learning could inspire new basic research in computer vision by making limitations of current approaches measurable and incentivizing the development of new methods to address these limitations. \new{The animal behavior community has already started facilitating such efforts by publishing datasets~\cite{sun2023mabe22, chen2023mammalnet, ma2023chimpact, fuchs2024forest, brookes2024panaf20k} and establishing benchmarks such as the Multi-Animal Behavior challenge \cite{sun2023mabe22} which uses data from lab experiments with mice, beetles, ants and flies.}

One of the biggest levers to enable progress in application domains such as animal behavior (but also more broadly) is to shift focus from developing architectures for existing tasks and datasets towards the \emph{process} of iteratively annotating (the right) data and training models until the accuracy requirements of the problem at hand are met. In other words: Given that we want to identify certain aspects of animal behavior (e.\,g. identities of individuals and grooming interactions), which parts (frames, snippets) of the dataset should we annotate, how (boxes, tracks, keypoints, actions spatially and/or temporally localized?) and in what order to achieve satisfactory performance with as little effort as possible? \new{This is the topic of data-centric learning. In computer vision, data-centric approaches are studied} mostly in the context of image classification, more recently also in object detection \cite{elezi2022not} and segmentation (e.\,g. SAM \cite{kirillov2023segment}). Extending such research more towards spatio-temporal tasks such as action and interaction recognition would be extremely helpful. To some extent, this is going to be a matter of adapting established \new{data-centric principles} to these more complex tasks and developing the right tools. However, there is also a deeper scientific question that has not received too much attention thus far: Action and interaction recognition is inherently a multi-task problem. Individuals need to be detected and tracked, then actions and interactions associated with them. Some of these tasks inform each other. For instance, it is well known that adding segmentation labels helps object detection (Mask R-CNN \cite{he2017mask}), but it is less clear whether investing the same amount of labeling time in additional bounding boxes would not improve a detector more. Similarly, some research has investigated how representations learned by one task generalize to another (Taskonomy \cite{zamir2018taskonomy}), but with a focus on purely spatial tasks, most of which are not directly relevant to the study of animal behavior. How much annotation effort should be put into each of the relevant tasks, and what are good strategies for identifying the limiting factors that provide the most leverage to improvement at any given point in time? This problem could be phrased as multi-task active learning, and we are not aware of any research in this direction.

\paragraph{Foundation models.}

\begin{figure}
    \centering
    \includegraphics[width=\textwidth]{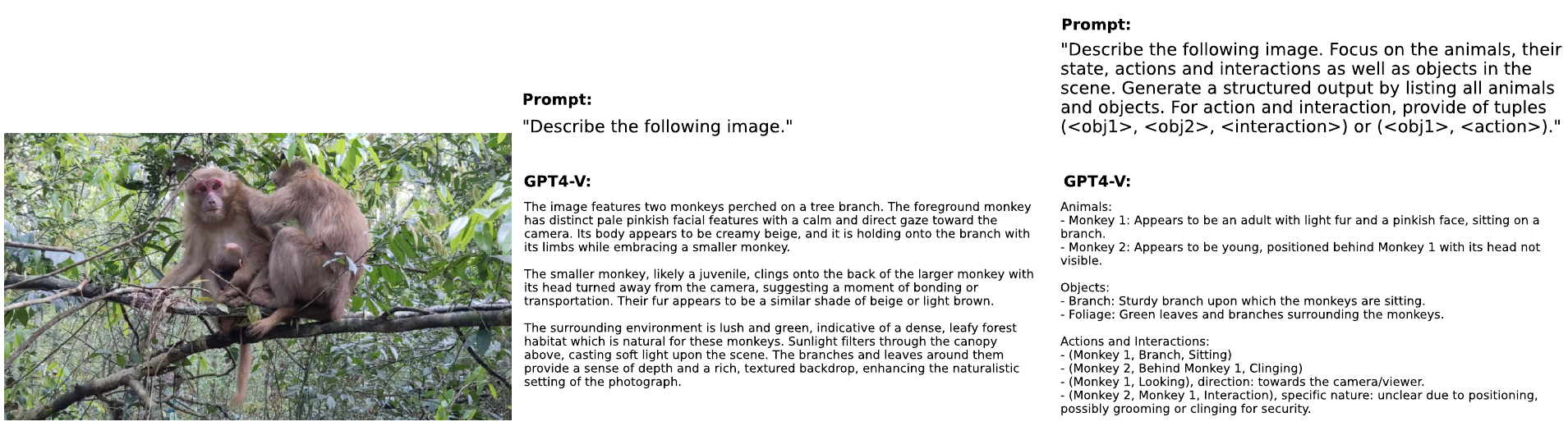}
    \caption{Multi-modal large language models (here GPT4-V) can be used to understand challenging scenes. It succeeds in recognizing the general setting (monkeys sitting on a branch) but fails to detect details (the baby monkey, spatial relation, interaction). While current models are not capable of a fully automated analysis they might serve as a tool for pseudo labeling. 
    More examples can be found in Appendix~\ref{sec:appGPT}.}
    \label{fig:gpt_single}
\end{figure}

ChatGPT has popularized the idea of foundation models~\cite{bommasani2021opportunities}, very large models that address several tasks at once. While the first versions operated exclusively on text, now image processing capability is available \new{in commercial as well as open source models such as LLaVA \newcite{liu2023llava} and Phi3-vision \newcite{phi3vision}}. This raises the question to which degree behavior analysis tasks will be solved by such general-purpose models in the future (Fig.~\ref{fig:gpt_single}). While early experiments \cite{yang2023dawn} suggest remarkable image understanding capabilities, we believe that \new{complete} behavioral analysis will not be conducted by these models in the near future for multiple reasons: (a) behavior requires video understanding and spatio-temporal reasoning and (b) the data domain is very different from typical internet videos. (c) Most importantly, tasks like individual identification, or specifying a target ethogram will necessarily require tuning the model as the required information to master these tasks is not accessible to the model.
Nonetheless, foundation models can already now enhance behavioral understanding: Even closed-source models such as OpenAI's GPT4-V can be leveraged to generate pseudo labels on images. From openly-available models such as SegmentAnything or CLIP, it is also possible to extract the knowledge through distillation \cite{Hinton2015DistillingTK}. 
\new{The recent AmadeusGPT \cite{ye2024amadeusgpt} and WildCLIP \cite{gabeff2024wildclip} showcase applications of foundation models for animal behavior understanding. AmadeusGPT answers Natural language questions about videos through Python code generation with GPT-3.5 and subsequent execution while using an API that involves DeepLabCut. WildCLIP builds on the open-vocabulary capability of the CLIP foundation model for camera-trap image retrieval.}

\section{Conclusion}

We gave a comprehensive overview of the current state-of-the-art computer vision techniques that we consider crucial for the analysis of individualized animal behavior in natural settings, with the four key tasks being object detection, multi-individual tracking, individual identification and action understanding. Additionally, we reviewed methods for effort-efficient learning, highlighting their practical significance given the extensive annotation requirements of computer vision models. 
We expect recent generic computer vision technologies, from multi-object tracking to action recognition, to surpass the performance of abundantly used specialized image- or keypoint-based methods for automated animal behavior analysis in the wild.

At the same time, there are still several computer vision problems highly relevant to practical behavioral researchers, which should -- in our opinion -- receive more attention from the community. Extending data-centric approaches from images to video and multi-task settings has great potential for primate behavior analysis.
Furthermore, unifying tasks such as tracking, individual identification and action understanding instead of treating them separately would enable a comprehensive characterization of how individual animals in a group move and what (inter)actions they engage in.
Creating such joint models is also desirable in terms of the convergence of different research directions towards holistic video processing. Ultimately, collaboration between behavioral scientists and computer vision researchers will play a vital role in finding the right path towards developing methodologies that are both powerful and practically applicable.

\section*{Acknowledgements}
This publication was funded by the Deutsche Forschungsgemeinschaft (DFG, German Research Foundation) -- Project-ID 454648639 -- SFB 1528 -- Cognition of Interaction and with NextGenerationEU funds from the European Union by the Federal Ministry of Education and Research under the funding code 16DKWN038. The responsibility for the content of this publication lies with the authors.

\newpage
\printbibliography[title={References}]
\end{refsection}

\newpage

\appendix

\begin{refsection}

\section{Foundations}
\label{sec:appFoundations}
        
\begin{figure}
    \centering
     \includegraphics[width=9cm]{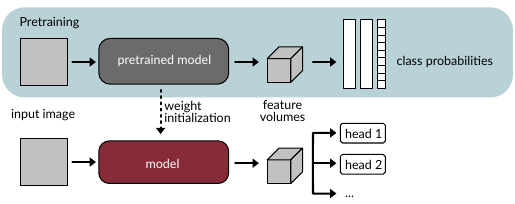} 
    \caption{Before fine-tuning for a specific task, often neural networks are initialized using ImageNet classification weights.}
    \label{fig:finetune}
\end{figure}

\subsection{Neural Networks}
Neural networks are machine learning models whose parameters (weights) can be efficiently learned from data using the backpropagation algorithm \cite{rumelhart1986learning}. Deep neural networks are composed of multiple layers, i.e. blocks of the conceptually same but differently parameterized operation. For instance, a fully-connected layer transforms an input vector by matrix multiplication. The transformed vector can then be used as input to another fully-connected layer.

\subsection{Backbones}
In modern neural-network-based computer vision, it is common to use a backbone model to obtain general image features, which can be applied to several tasks. Backbones incorporate best practices for the design of the neural network architecture such that this does not have to be reinvented for every task.
Convolutional neural networks (CNNs) such as ResNets \cite{he2016deep} are commonly used as backbones for image tasks. The idea here is to apply the same set of learnable filters at a grid of locations of an image in each layer. Since the same operation is repeated at different locations and the number of parameters is independent of the image size, CNNs are parameter-efficient models that can learn from small datasets.
The feature pyramid network \cite{Lin2016FeaturePN} is a CNN that can deal with image content at different size scales.
Driven by the success in language modeling, transformer models were recently successfully applied on images \cite{dosovitskiy2020image}. The input image is divided into a grid of patches (commonly of size 8 to 16 pixels). Each patch is mapped to a token using a linear projection and encoded using a positional embedding. Then, a transformer encoder \cite{Vaswani2017AttentionIA} is used to process the image. To extract a single feature vector for whole-image classification, a CLS token can be added to the input (in addition to the patches) or all tokens can be pooled.

The choice of backbone strongly influences training and inference time, as the number of parameters and compute operations can vary substantially. Backbones with a large number of parameters tend to achieve better performance across downstream tasks but require more time to train and to make predictions at test time. Such backbones might also not fit onto embedded devices to be used in the field (e.\,g. a Raspberry Pi 4) or might not permit higher frame rates to study motion in real-time.

\subsection{Pre-training and Transfer Learning}
Pre-training (Fig.~\ref{fig:finetune}) describes the process of training a model on an external, often large, dataset, which allows the model to learn generic image or video features that represent the visual content of the input. During transfer learning, this pre-trained model is fine-tuned on a, typically smaller, dataset that is available for the task of interest. For example, an object detection model that has been trained on a large dataset that consists of humans can be used for fine-tuning on a smaller object detection dataset that consists of primates. We provide more detail on transfer learning and different pre-training techniques in Appendix~\ref{app:trans}.

\subsection{Task Heads}
Starting from a pre-trained backbone, the desired task can be trained as a task head on top of it. For example, if our target task is individual identification, we can add an ``identification'' head to the backbone.  Multiple task heads may be added to the same backbone (e.\,g. identification, localisation,  action classification). Since we expect the backbone to have learned good features before, the target task does not need a very large number of parameters. If the downstream dataset is small and the pre-trained backbone has many learnable parameters, fine-tuning the entire model can lead to overfitting. This means that the model performs well on the training set, but loses its ability to generalize to unseen data. Therefore, it can be helpful to freeze most layers and only train a small part of the model, such as the task heads.

\section{Methods for Primate Behavior Analysis}
\label{sec:appCV}

\subsection{Animal detection}
\label{app:detect}

The goal of detection is to take an image and predict a set of bounding boxes that indicate the locations of animals in the image. Often the bounding box locations are accompanied by object class predictions and confidence scores.

\paragraph{Anchor-based}
One approach for object detection is to define a large set of candidate regions, called anchors, a priori and allow the model to determine which of these candidate regions matches an object. This means the task of determining the presence of an object is reduced to predicting a confidence score that indicates if each candidate region contains an object or not. However, since the number of all possible object regions is very high, it is computationally infeasible to make a prediction for all of them. Many anchor-based detection methods solve this problem using a sliding-window approach \cite{Ren2015FasterRT, liu2016ssd, Redmon2018YOLOv3AI}: They pre-define a small number of anchor-boxes (usually not more than 10) by their width and height to cover objects of different scales. Then, the image is divided into a grid and for each grid location and anchor box, it is evaluated whether there is a matching object or not. For computational efficiency, the grid does not cover every pixel. To compensate for the inaccuracies introduced by considering only a fixed set of anchor boxes and grid locations, an offset is predicted in addition to the confidence score to refine the location as well as the width and height of the detected objects. To obtain a set of problem-specific anchor-box widths and heights, their distribution in the training dataset can be used, e.\,g. a pedestrian dataset will result in primarily elongated anchors. A disadvantage of anchor-based approaches is that the model's performance might decline if the bounding boxes in the target dataset look different from the training distribution. Notable examples of single- and two-stage methods employing anchors are the R-CNN \cite{Ren2015FasterRT, He2017MaskR} and YOLO family \cite{Redmon2016YOLO9000BF, Redmon2018YOLOv3AI}.

\paragraph{Heatmap-based}
Heatmap-based detection is an object detection paradigm that does not require anchor boxes. Here, a heatmap (a one-channel feature map) is generated over the entire image indicating the predicted locations of the corners or center of each object. The predicted values of this heatmap should be high at specified object locations, such as the object center or object corners, and zero everywhere else. 

In CornerNet \cite{Law2018CornerNetDO}, top-left and bottom-right pairs of keypoints indicating bounding box locations are predicted as two sets of heatmaps (one per category, such as monkey or human, to enable multi-class detection). To resolve the ambiguous correspondence between pairs of keypoints when multiple objects of the same class are present, an additional pixel-wise embedding is trained to be similar at locations of corresponding points.
CenterNet (by Duan~et~al. \cite{Duan2019CenterNetKT}) extends this approach by predicting the center point of the object in addition to the two corner keypoints.
Another method also called CenterNet (by Zhou~et~al. \cite{Zhou2019ObjectsAP}) predicts the centerpoint of each object as a heatmap (one per category) and generates bounding box width and height in a different prediction head. The center heatmap indicates at which position the width and height information should be read out.
The Fully-Convolutional One-Stage Object Detector (FCOS \cite{tian2019fcos}) employs a similar approach to the heatmap-based approaches but uses all pixels that pertain to an object to estimate the location and size of its bounding box instead of just considering the center or corners.

\paragraph{Detection transformer: DETR}

The output of many of the previous detection methods cannot be directly used but needs to be further processed to avoid predicting the same bounding box multiple times at nearby locations. This heuristic filtering of redundant bounding boxes is a non-differentiable operation and, depending on where in the framework it is employed, it might prevent the network from being trainable in an end-to-end manner \cite{Ren2015FasterRT}. This complicates the training procedure and can have a negative impact on overall detection performance \cite{sun2021makes}. The detection transformer (DETR) \cite{Carion2020EndtoEndOD} introduces the concept of object queries, which can be used to make unique object predictions without the need for heuristic filtering or other hand-crafted components, such as anchors. Each object query is a predefined or learned numerical vector. At its core, DETR relies on a transformer that decodes these object queries into bounding boxes using image features (specifically, a high-level feature map) from a CNN backbone. Since information exchange between object queries is possible through self-attention, duplicate matches to ground truth objects can be prevented by penalizing them in the training objective. 

In practice, DETR has slow convergence (i.e. requires many training iterations) and high computational demands due to the quadratic complexity of the transformer. Subsequent works aimed to tackle these issues \cite{Zhu2020DeformableDD, Li2022DNDETRAD, Liu2022DABDETRDA}. For example, in Deformable-DETR \cite{Zhu2020DeformableDD} each object query learns to attend to a small number of locations that are relevant for object detection, which reduces computational complexity and leads to faster convergence. Various improvements of DETR have been combined, resulting in the DINO detector\footnote{this should not be confused with self-supervised learning method DINO} \cite{Zhang2022DINODW}, which is currently the state of the art in object detection.

\paragraph{Practical Detector Training Considerations}
Besides the model's architecture, i.e. backbone and detection method, the training algorithm, training data and loss function are crucial factors for detection performance. Over the years, multiple techniques have been developed. 
For example, one common problem of object detection datasets is class imbalance, i.e., some object classes occur much more frequently than others. This imbalance incentivizes a model to improve on frequent rather than rare classes during training. Focal loss \cite{Lin2017FocalLF} adds a factor to each class term in the loss function, such that classes that are not yet classified that well receive more weight. Augmentations, i.e. random modifications to the images and labels, increase the variability of the training data and have emerged as a crucial component in object detection. Classic augmentations include random cropping and change of contrast. More recently, mosaic augmentation and mixup proved successful in the context of object detection \cite{Ge2021YOLOXEY}. Another recent advancement is SiMoTA \cite{Ge2021YOLOXEY}, an algorithm that seeks to optimize which predictions are considered positive and negative during training. Using this advanced label assignment strategy has a positive impact on both training time and model performance.

\paragraph{Datasets and Benchmarks}
The standard benchmark for evaluating an object detection model is the Common Objects in Context (COCO) dataset \cite{Lin2014MicrosoftCC}. Since virtually all published detection methods provide metrics for their performance on COCO, it is a useful benchmark to help researchers select the right model in terms of accuracy and computational efficiency. However, model performance on COCO might not be directly comparable to performance on tasks involving the detection and individual identification of animals, as COCO contains a diverse set of 80 classes, and is more suited for broader classification problems (e.\,g. classifying an object as a dog or a bicycle, rather than distinguishing between similarly looking individuals of the same class, as required for primate detection tasks). 
The Objects365 \cite{Shao2019Objects365AL} and LVIS \cite{Gupta2019LVISAD} datasets include ``monkey'' as an object class, although in both cases, this class includes a wide variety of taxonomic groups, for example platyrrhini, catarrhini (both cercopithecoids and apes), and prosimians, as well as artificial objects, such as toys, that resemble primates.

\new{Furthermore, there are several datasets with primates for keypoint and pose estimation. The MacaquePose dataset \cite{Labuguen2020MacaquePoseAN} contains more than \num[group-digits=integer]{13000} images of macaques with labeled keypoints.
OpenApePose \cite{Desai2022OpenApePoseAD} offers over \num[group-digits=integer]{70000} images of six ape species with annotated poses. The AP-10K \cite{yu2021ap} and APT-36K \cite{yang2022apt36k} datasets are part of the larger Quadruped-80K dataset for pose estimation of several animal species, including monkeys \cite{ye2022superanimal}. Other datasets for pose estimation include OpenMonkeyChallenge \cite{yao2023openmonkeychallenge} with over \num[group-digits=integer]{111000} annotated frames of 26 primate species in the wild, and APT-v2 \cite{yang2023aptv2} pose tracking for many animal species on short video clips, including gorillas, chimpanzees, and orangutans.}

\subsection{Multi-animal tracking}

\paragraph{Tracking-by-detection}
Most work in the tracking-by-detection paradigm is based on the seminal SORT model \cite{bewley2016simple}. In this model, detections of each frame are collected independently of each other using Faster R-CNN \cite{Ren2015FasterRT}. The spatial shift between detections of successive frames is then used to extract motion features. Next, the Kalman filter algorithm \cite{kalman1960new} is applied which uses these motion features and the previous location of an object to determine the predicted bounding box on the next frame. The overlap between the predicted and detected bounding boxes is used to compute the affinity cost for all possible pairings. Finally, the model associates the detections across frames by minimizing the affinity cost using Hungarian matching \cite{kuhn1955hungarian}. This association process is continued for each frame of the video to obtain tracks.

Within the SORT framework, a number of models have emerged, each tweaking different components of the model to achieve superior results. Most models now implement more powerful backbones, such as Darknet \cite{zhang2022bytetrack,cao2022observation}, to facilitate superior feature extraction. To improve association, models often incorporate appearance characteristics of detections \cite{zhang2022bytetrack,cao2022observation,maggiolino2023deep}. Kalman filters have also seen improvements \cite{cao2022observation,du2023strongsort,maggiolino2023deep}, and adaptive thresholds for pairing detections have been established \cite{zhang2022bytetrack,aharon2022bot}. Some models account for camera motion \cite{aharon2022bot} and have made modifications to the overlap computation for two bounding boxes\cite{yang2023hard,yan2022multiple}. In other works, affinity costs are computed indirectly by a graph neural network based on all detections over multiple consecutive frames \cite{braso2020learning,Cetintas2022UnifyingSA}.

\paragraph{Tracking-by-query}
Apart from the classical tracking-by-detection framework, there are also methods with information exchange between the detection and association phase. This information exchange has been facilitated by the advent of transformer structures in computer vision. Specifically with the introduction of DETR \cite{Carion2020EndtoEndOD} in object detection, the number of models implementing this tracking approach has seen a significant increase \cite{zeng2022motr,Meinhardt2021TrackFormerMT,zhan2022robust, sun2020transtrack}. When using the DETR architecture (see Section~\ref{sec:object_detection}) for tracking, the resulting embeddings from the transformer decoder are not only mapped onto bounding boxes or masks. Rather, they are also reused as additional track queries, which guide the detection and association process in the following frame. This way, motion and appearance information from previous frames is taken into account for both detection and association. Significant progress in this area has been made with MOTRv2 \cite{Zhang2022MOTRv2BE}, which uses external detections from the YOLOX detection model \cite{Ge2021YOLOXEY}, that are associated and potentially refined using track queries. Additionally, an improved variant of DETR \cite{Liu2022DABDETRDA} contributes to enhanced processing. Most recent works even eliminated the need for an external object detector in this architecture by employing a specialized training strategy that reconciles the detection and association objective \cite{yu2023motrv3, yan2023bridging}.

\paragraph{Datasets and Benchmarks}
Since most tracking projects are interested in tracking humans, they are typically evaluated on human multi-object tracking benchmarks, such as MOT17 \cite{milan2016mot16}, a dataset of annotated videos of pedestrians walking through the streets, or DanceTrack \cite{sun2022dancetrack}, which contains annotated videos that show dancers and, therefore, humans engaged in more complex, non-linear motion patterns than in MOT17. Furthermore, the dancers in a video tend to be similarly clothed, which makes the tracking task even harder. Models performing well on DanceTrack may be good candidates for animal tracking tasks because in this domain we also expect similar appearances and highly non-linear motion patterns. Recently, the first dataset solely devoted to tracking groups of animals was introduced (AnimalTrack, \cite{zhang2023animaltrack}). Primate videos are not included in this dataset.

\subsection{Individual identification}

Training a classification or metric-based model for individual identification requires annotated images or videos of animals alone or in a group. Annotation involves the localization of an animal in a frame, e.\,g. by drawing a bounding box, and then assigning an identity to it. Previous studies have demonstrated promising identification results with annotated training sets containing merely tens to hundreds of examples \cite{reid_face2, korschens2018towards, moskvyak2021robust}, while other works have used thousands of training examples per individual \cite{reid_bird, romero2019idtracker}. The exact number of required annotated examples depends on the complexity of the task and the desired model accuracy. When applying identification models to unseen data, it can be challenging to recognize individuals in single frames, due to occlusion or non-standard poses. Therefore, researchers often utilize camera traps which produce bursts of several images taken in quick succession when they detect motion \cite{cam_trap}.  When several images from a sequence are available, the model predictions can be aggregated, which can lead to improved performance compared to when model predictions are based on single images  \cite{reid_svm}. This aggregation technique can be combined with multi-object tracking methods, so that the model only has to be very certain on a few frames to propagate the identification result to all other frames of the video.

\paragraph{Datasets and Benchmarks}
\new{Individual identification research is primarily driven by human-centered benchmarks. There are several datasets for face recognition \cite{whitelam2017iarpa, maze2018iarpa, kalka2018ijb, cheng2019low, guo2016ms, huang2008labeled} and recognition based on the full body \cite{zheng2016mars, zheng2015scalable}}

\subsection{Action understanding} \label{app:action}

\paragraph{Action recognition}

Action recognition models classify what action a video sequence contains. Early action recognition methods primarily rely on 3D-convolutions on RGB inputs \cite{qiu2017learning, lee2021diverse, feichtenhofer2019slowfast}. For example, SlowFast \cite{feichtenhofer2019slowfast} convolves over low and high frame rate input to capture high-level semantics and fine motions. The main drawback of convolutional networks is their limited receptive field which is of particular disadvantage in video-understanding tasks that often require long-range temporal reasoning. Therefore, there has been a shift towards transformer networks which benefit from stronger temporal modeling capabilities. The pioneering works in this area \cite{arnab2021vivit, bertasius2021space} closely follow the design of Vision Transformer \cite{dosovitskiy2020image}. To tackle the additional computational cost imposed by the time dimension, the authors proposed several methods to minimize the memory load, such as tube embedding, frame subsampling and various attention factorizations. Subsequent efforts extended these transformer models by including feature pyramids \cite{yan2022multiview, fan2021multiscale, ryali2023hiera}, by enhancing their long-term modeling capabilities \cite{wu2022memvit} or by improving efficiency \cite{bulat2021space, xiang2022spatiotemporal}. Some works also combined the merits of convolution and transformer models by modeling local and global features in shallow and deep layers respectively \cite{li2022uniformer, li2022uniformerv2}.

Another line of research approaches action recognition by modeling motion-related features. Early works incorporated optical flow features in convolutional architectures \cite{tran2015learning, carreira2017quo, xie2018rethinking} while more recently there have been attempts to design neural network modules to compute motion features \cite{zhang2020pan, kwon2020motionsqueeze}. Furthermore, there are also methods which use motion information to guide feature extraction \cite{zhao2018trajectory, patrick2021keeping, long2022stand, wang2021tdn, kwon2021learning, yang2020temporal, liu2022motion, huang2022busy}. For example, efforts have been made to aggregate features along motion trajectories \cite{zhao2018trajectory, patrick2021keeping, long2022stand}, which accounts for the fact that the same object may appear in different locations in different frames across time due to the movement of the object. Another strand of literature models interactions between objects \cite{wang2018videos, zhou2018temporal, materzynska2020something, tian2022ean}. For example, the action ``feeding from bowl'' can benefit from explicitly modeling the interaction between the feeding individual and the bowl. While early approaches rely on external object detectors \cite{wang2018videos, materzynska2020something}, more recent frameworks leverage attention to model interactions implicitly \cite{tian2022ean}. Although the modeling of motion and action-related features provides suitable inductive biases for action recognition, performance often falls short compared to the more generic transformer-based video backbones. This is especially true if these backbones are pre-trained with state-of-the-art self-supervised methods \cite{wang2022internvideo, tong2022videomae}. Nevertheless, the exploration of inductive biases for video-understanding tasks remains an important research topic since they can be helpful in ensuring a decent model performance in the face of limited training data. In contrast, generic video backbones usually require huge amounts of labeled data when being trained from scratch.

There are several human-centered action recognition benchmarks \cite{sigurdsson2016hollywood, damen2020rescaling, kay2017kinetics, soomro2012ucf101, kuehne2011hmdb, goyal2017something, monfort2019moments}. These benchmarks consist of short videos (a couple of seconds) that are labeled with one or more action labels. The Animal Kingdom dataset \cite{ng2022animal} contains video-level action labels for a diverse set of animal species, but does not include primate videos.

\paragraph{Temporal action detection}

Temporal action detection methods assign action labels to temporal intervals in a video. They can be divided into single- and two-stage methods, analogous to detection methods. Two-stage methods first generate temporal proposals which are subsequently classified and potentially refined. Single-stage methods directly produce predictions without the in-between proposal generation step. 

Early proposal generation methods are based on anchors, i.e. multiscale temporal intervals centered at various temporal locations \cite{shou2016temporal, xu2017r}. For example, \cite{shou2016temporal} obtain proposals by classifying anchors as containing or not containing an action based on the features associated to the anchor. Since anchors might be too restrictive to capture the large temporal variety in ground truth action instances, subsequent works developed anchor-free proposal generation methods. In their pioneering work, \cite{Lin2018BSNBS} predict starting, ending and action probabilities at individual temporal locations and arrive at proposals by grouping according to these scores. Empirically, grouping based on low-level predictions can be unreliable, for example due to scene switches. This issue can be counteracted by enhancing the temporal context modeling capabilities or employing regularization techniques, which has been pursued in subsequent works \cite{Su2020BSNCB, Lin2019BMNBN, Lin2019FastLO, Song2023FasterLO, Chang2021AugmentedTW, Zhao2020BottomUpTA}. For example, \cite{Chang2021AugmentedTW} employ a graph-convolutional network and a transformer to model local and global relationships between video snippets, respectively. \cite{Zhao2020BottomUpTA} regularize proposals by enforcing consistency between and within starting, actionness and ending phases. Methods that take proposals as input either operate directly on proposal-level features \cite{Wang2017UntrimmedNetsFW} or further enhance them by proposal-context modeling \cite{Shou2017CDCCN, Zeng2019GraphCN, Zhu2021EnrichingLA} to produce the final predictions.

Early single-stage methods were also based on anchors \cite{lin2017single, liu2020progressive}. For example, the first work to propose single-stage processing \cite{lin2017single} directly regressed action boundaries and class confidence from anchors similar to SSD for object detection \cite{liu2016ssd}. More recently, anchor-free methods have demonstrated superior performance \cite{long2019gaussian, lin2021learning, liu2022end, he2022glformer, cheng2022tallformer, zhang2022actionformer, tang2023temporalmaxer, shi2023tridet}. In this line of research, attention-based methods that operate on snippet-level features computed by pre-trained 3D-convolutional networks are a popular approach. For example, ActionFormer \cite{zhang2022actionformer} successively employs transformer and subsampling blocks to snippet-level features to create a feature pyramid, which facilitates the detection of actions at different temporal scales. The resulting tokens are used to regress the relative offset to the action boundaries and class confidences. Recently, transformers have been criticized for their inability to attend to crucial frames among frames with similar appearance \cite{tang2023temporalmaxer}. More recent approaches have therefore replaced attention with simpler temporal context modeling techniques while achieving state-of-the-art results \cite{tang2023temporalmaxer, shi2023tridet}. For example, TemporalMaxer \cite{tang2023temporalmaxer} follows the same design as ActionFormer but replaces the transformer layers with simple max-pooling operations. Remarkably, these approaches even achieve state-of-the-art performance in dense prediction tasks where there are multiple overlapping actions, outperforming methods specifically developed for these tasks \cite{dai2021ctrn, dai2021pdan, tirupattur2021modeling, dai2022ms, tan2022pointtad}. 

There are various benchmarks for regular \cite{idrees2017thumos, heilbron2015activitynet} and dense \cite{yeung2018every, dai2022toyota, sigurdsson2016hollywood, damen2020rescaling} temporal action detection.

\paragraph{Spatio-temporal action detection}

Existing spatio-temporal action detection methods can be divided into two categories. First, there are track-based methods that identify tracks of actors in a video and assign action labels to sub-tracks. Second, there are frame-based methods that leverage spatio-temporal information to assign one or more actions to detections in individual video frames.

Early methods for track-based spatio-temporal action detection usually employed 2D-convolutions on frame-level RGB inputs and optical flow features to obtain action localizations, which are then linked via tracking algorithms \cite{weinzaepfel2015learning, gkioxari2015finding, saha2016deep}. Subsequent works improved upon this approach by leveraging temporal information \cite{kalogeiton2017action, yang2019step, li2020actions, zhao2022tuber, singh2023spatio, gritsenko2023end}. This is done by directly predicting action tracks on short video segments and subsequently linking them based on overlap to obtain video-level action tracks. For example, this paradigm was adopted by STAR \cite{gritsenko2023end}, which uses a set of spatio-temporal queries to decode action tracks from segment-level video features. Notably, the method can be supervised with action tracks sparsely annotated on individual keyframes, which are easier to obtain than exhaustive annotations. Another recent work \cite{rajasegaran2023benefits} represents detections in individual frames by their appearance and estimated 3D pose features. All detections in a video segment are then input to a standard transformer architecture to predict framewise action classes which are linked with a 3D tracking algorithm.

In contrast to track-based methods, frame-based spatio-temporal action detection aims to detect actions on individual keyframes. A separate tracking algorithm is needed to link these detections into action tracks. The standard frame-based approach \cite{gu2018ava} can be divided into three steps. First, a pre-trained object detector like faster R-CNN \cite{jiang2017face} is used to obtain bounding boxes of actors in a frame. Then, each bounding box is symmetrically extended along the temporal dimension to obtain a 3D region of interest, which is used for feature extraction from a video backbone. Lastly, the extracted features are used for multi-class action classification. This procedure has been employed in combination with various video backbones \cite{feichtenhofer2019slowfast, feichtenhofer2020x3d, dosovitskiy2020image, fan2021multiscale, wu2022memvit} and self-supervised pre-training \cite{wang2022internvideo, tong2022videomae}.

However, there have also been efforts to develop more sophisticated modeling frameworks for frame-based spatio-temporal action detection. Early approaches focused on modeling actor-context \cite{sun2018actor, girdhar2019video} or long-term temporal \cite{wu2019long} relations. To illustrate why it is useful to model these two types of relations, consider an individual that is feeding. In this case, contextual cues, such as the proximity of a feeding bowl, or temporal information, such as the individual taking the food a couple of seconds before, are indicative of the performed action. Subsequent research acknowledged that both context and temporal relations must be modeled simultaneously to make further progress \cite{tang2020asynchronous, arnab2021unified}. For example, \cite{arnab2021unified} construct a spatio-temporal graph with actor and context detections from successive video snippets and employ attention-based message passing. The enriched actor-features are then used for action-classification. Other works include more intricate relationships in their models, while employing a memory module for temporal reasoning. For example, ACAR-Net \cite{pan2021actor} includes higher-order interactions between actors and contextual objects and HIT \cite{faure2023holistic} uses both RGB and pose features to model hand, object and person interactions. Notably, all of these more sophisticated modeling attempts perform worse than the standard three-step frame-based approach with transformer-based backbones and self-supervised pre-training.

Benchmarks for track-based spatio-temporal action detection, include the \mbox{J-HMDB-21} \cite{jhuang2013towards}, \mbox{UCF101-24} \cite{soomro2012ucf101}, MultiSports \cite{li2021multisports} and AVA datasets \cite{gu2018ava, li2020ava}. However, the AVA dataset only contains sparsely annotated action tracks at one frame per second and, therefore, has been primarily used for the evaluation of frame-based approaches. 

\paragraph{Dynamic scene graph generation}

Methods for dynamic scene graph generation (DSGG) localize animals and objects and predict relationship triplets of the form entity--relation--entity, where entities can be animals or objects. There are two paradigms for DSGG: In track-based methods, entities are tracked over time and relations are predicted for sub-tracks of two entities. In frame-based methods, relations are only predicted for individual frames of a video. It should be noted that there is also an extensive literature on interaction detection and scene graph generation methods for still images, however, since most relations and actions can be more easily identified when temporal information is available, we focus on video-based methods in this work and refer to recent survey papers for an overview of image-based approaches \cite{antoun2022human, zhu2022scene}.

Irrespective of the paradigm, all DSGG methods roughly follow the same procedure: First, features are extracted for entities and relations. The entity features are usually obtained from image- or video-backbones, while relations are often represented by the class labels, union features (i.e. features of both involved entities) and relative motion of subjects and objects. In a second step, the extracted features are further processed using some form of spatio-temporal reasoning. Finally, the features are used to separately predict the entity and relation classes. A joint prediction of the relationship-triplet is avoided because the number of possible classes would be intractable.

Early track-based methods divide the input video into short, overlapping segments and generate scene graphs for each segment separately \cite{qian2019video, tsai2019video, shang2017video}. Video-level scene graphs are then obtained by merging segment-level relationship-triplets if the corresponding tracks have a sufficient overlap and the predicted relation is the same. Many subsequent works generally maintained this paradigm, while trying to improve on individual aspects \cite{sun2019video, su2020video, li2021interventional, shang2021video}. For example, \cite{su2020video} realized that the proposed merging algorithm is not robust to segment-level errors. Therefore, they proposed to preserve multiple relationship-triplet hypotheses at each merging step and discard unlikely ones when sufficient information is available. However, the problem remains that scene graphs are generated separately for each segment without using the information from other segments, which prevents the learning of inter-segment dependencies. For example, there is often a natural sequence of relations (e.\,g. person--take--cup followed by person--look at--cup). This problem has been addressed by \cite{wei2023defense} who used two separate graph-convolution-based modules for segment-level inter-entity modeling and video-level inter-relation modeling to achieve state-of-the-art results for track-based DSGG. In addition to this work, there have also been efforts to discard the segment-level paradigm altogether \cite{liu2020beyond, gao2022classification, chen2021social, gao2021video}. For example, \cite{gao2021video} perform tracking on the complete input video and obtain sub-tracks using a sliding window approach with multiple kernel sizes. Then, a pair proposal module is used to identify pairs of sub-tracks that relate to each other. In the last step, the entity and relation classes of the identified pairs are predicted. Although methods that operate on complete videos facilitate global processing, they have problems of their own. For example, the creation of long tracks is prone to errors, which can have a negative impact on DSGG performance. In addition, the features of long tracks must be heavily compressed to be processable, resulting in a loss of information.

Frame-based methods are dominated by transformer models. In early approaches, the dependency between all relation or entity features of neighboring frames is modeled \cite{ji2021detecting, arnab2021unified, cong2021spatial}. For example, \cite{cong2021spatial} use two separate but identical transformer architectures for intra- and inter-frame relation-modeling. Although this approach is refreshingly simple, it suffers from memory issues since the temporal reasoning is done in a fully-connected way. Methods along this line are therefore restricted to operating on a small number of frames. In contrast, more recent methods rely on some form of tracking to perform temporal modeling for each subject-object pair separately \cite{wang2023cross, li2022dynamic, feng2023exploiting}, thus allowing for the consideration of a broader temporal context.

Benchmarks for track-based DSGG include the VidVRD \cite{shang2017video} and the VidOR \cite{shang2019annotating} datasets, while for frame-based DSGG there is the Action Genome dataset \cite{ji2020action}. All of these benchmark datasets consist of videos containing various human-centered relations. These relations include actions, but also spatial arrangements. We are not aware of any animal related benchmarks.

\paragraph{Memory limitations}

\new{
Although action understanding frameworks are largely based on video-level processing, these models are not widely used in primate behavior research. This is partly due to the higher memory requirements of video-based models compared to their image-based counterparts. However, we argue that these constraints will be overcome in the near future. Firstly, current trends in hardware development suggest that GPUs with a high memory capacity will become readily available to a broader audience. Secondly, computational efficiency of methods is an active area of research such that powerful video-based methods are becoming more and more compute efficient \cite[see e.\,g., ][]{ryali2023hiera}. Already today, popular video backbones used in state-of-the-art action understanding methods can be trained with reasonably large batch sizes on a single GPU. For example, MVD \cite{wang2023masked}, which uses VIT \cite{dosovitskiy2020image} as a backbone, requires only about 1.5~GB of VRAM per input clip for training the base model with 87~M parameters. Similarly, MVIT \cite{fan2021multiscale} and Hiera \cite{ryali2023hiera} require less than 2~GB VRAM per input clip for training the base model with 36.5~M parameters. The standard input clip consists of 16 images with a resolution of $\text{224}\times\text{224}$ pixels. Since the model inference consumes even less memory, applying these models on consumer-grade GPUs is not a problem. Therefore, we argue that, even today, the use of video-based models for action understanding still has unexplored potential. }

\section{Methods for effort-efficient learning}
\label{sec:appEffortEff}

\subsection{Transfer Learning} \label{app:trans}
Transfer learning involves leveraging knowledge gained from one task to improve learning and performance on a related, but different, task. By reusing previously acquired features or models, transfer learning addresses data scarcity and accelerates model training. This approach is particularly valuable in scenarios where labeled data for the target task is limited or expensive to obtain.

Traditionally, models were pre-trained through supervised learning on source tasks with labeled data \cite{ sharif2014cnn, talukdar2018transfer, Radford2021LearningTV, ye2022superanimal}, transferring knowledge to target tasks with limited labeled data. In the realm of computer vision, a prevalent transfer learning approach involves pre-training on an extensive and diverse dataset such as ImageNet \cite{Russakovsky2014ImageNetLS}, MS-COCO  \cite{Lin2014MicrosoftCC}, or Open Images Dataset \cite{kuznetsova2020open}, followed by fine-tuning on a narrower target dataset. However, recent advancements have introduced the use of unlabeled data as a pivotal paradigm for pre-training in transfer learning \cite{chen2020simple, chen2021exploring, zbontar2021barlow, Oquab2023DINOv2LR}. These methods allow models to be pre-trained on large amounts of unlabeled data by predicting or reconstructing certain data aspects, eliminating the need for explicit labels. A more elaborate discussion of these self-supervised learning methods is presented in Appendix~\ref{app:selfsuper}.

Modern successors of transfer learning have reached zero-shot transfer capabilities, which means that no examples are required to fine-tune the model to the target task. Instead, the diversely pre-trained foundation model is able to adapt to a target task directly. Language-vision models like CLIP \cite{Radford2021LearningTV}, EVA \cite{fang2023eva}, Florence \cite{yuan2021florence} and Flamingo \cite{alayrac2022flamingo} are big models which are trained on large-scale datasets, usually involving text and associated images obtained from the internet. 
These models have a good general understanding of many object classes and often achieve a zero-shot performance that is on par or better than smaller models trained specifically for this task. An advantage of these models is their strong ability to use information encoded in natural language, so that complex questions about an image or video could be formulated as a text, lowering the technical requirements on the user side for working with such a system.

\subsection{Self-supervised learning} \label{app:selfsuper}
\label{sec:self-supervised}

The objective of self-supervised learning is to derive labels directly from the data, enabling the model to learn meaningful representations from large, unlabeled datasets. 
These representations serve as valuable resources for training general-purpose models that can be applied to a wide range of downstream tasks. Therefore, self-supervised learning techniques produce models which can be used as foundations for transfer learning. In the following paragraphs, we delve into several state-of-the-art approaches in self-supervised learning, which can be broadly classified into three categories: pretext task learning, joint embedding architectures, and generative approaches. 

\emph{Pretext task learning} trains models on tasks like predicting image patch positions \cite{doersch2015unsupervised} or image rotations \cite{gidaris2018unsupervised} to learn data features without labels. These tasks promote understanding of object features and spatial relationships but pose challenges in task design and potential over-specialization that may impair future performance on tasks such as classification or regression.

In \emph{joint embedding architectures}, representations are learned by minimizing the deviation between similar samples and maximizing the deviation between dissimilar samples provided as training data. Typically, a contrastive loss is optimized between augmented pairs of similar and dissimilar samples. 
The SimCLR \cite{chen2020simple} framework is a popular contrastive learning framework that led to the development of many succeeding contrastive learning approaches based on a similar idea. The main idea of the framework is to maximize similarity between different versions of similar samples (created through augmentation techniques like cropping and rotating), while minimizing the similarity between the dissimilar sample pairs. 
However, this contrastive self-supervised learning method requires a large number of dissimilar examples for training to prevent collapse of the representation into trivial solutions, where all inputs map to the same output, and it requires substantial batch sizes for optimal performance.
Recent studies, including SimSiam \cite{chen2021exploring}, BYOL \cite{grill2020bootstrap}, Barlow Twins \cite{zbontar2021barlow} and DINO \cite{caron2021emerging} have made significant advancements in the field of self-supervised learning by eliminating the need for dissimilar or negative samples. These methods employ various techniques, such as momentum encoder and cross-batch memory banks, to learn representations without explicitly utilizing negative pairs. VicReg \cite{bardes2021vicreg} addresses the issue of collapsing by introducing a regularization term that focuses on the variance of the embeddings. Notably, these techniques have exhibited competitive performance when compared to SimCLR across diverse benchmark datasets. 

A noteworthy \emph{generative method}, known as Masked Autoencoders (MAE) \cite{he2022masked}, has shown immense promise in the realm of self-supervised learning. The fundamental concept underlying MAE involves masking random patches within an input image and subsequently reconstructing the missing pixels. Building upon the success of this approach in image-based downstream tasks, researchers have extended this idea to video-based pre-training \cite{tong2022videomae,wang2023videomae}.
A combination of masked image modeling and contrastive learning is found in iBOT \cite{zhou2021ibot} and CMAE \cite{huang2023contrastive}.

\subsection{Weakly supervised learning}

Perfect annotations are beneficial for model training, but not always available. In general, imperfections can have several origins, for example, bad annotation quality, partly wrong annotations, or labels that were produced for a similar but not exactly the desired task among others. The goal of weakly supervised learning is to find an effective way to extract information from such imperfect data to solve the desired task. Weakly supervised learning has been used in a wide array of computer vision tasks, including object localization \cite{shao2022deep}, image classification \cite{mahajan2018exploring}, pose estimation \cite{zhou2017towards}, action localization \cite{lee2020background} and scene understanding \cite{shi2021simple}. 

Given the extensive annotation effort necessary in the context individualized behavioral analysis, the most important application of weakly supervised learning involves the use of coarse-grained labels for a target task with a finer granularity. For example, the goal might be to train a model for object localization using only information about how many individuals are present in an image. Current methods used for coarse-to-fine-grained weakly supervised learning are mostly based on one of two paradigms: Multiple Instance Learning (MIL) and Class Activation Maps (CAM).

In \emph{MIL-based} approaches \cite{ren2020instance, wan2019c} several images (or frames of a video) \{$X_1,...,X_n$\} are combined into bags. Labels must only be provided at the bag level, i.e. when any of the images contain the labeled object, the complete respective bag receives this label. For example, the label ``dog'' would be given to the bag if the bag contains at least one image with a dog. The models have to find out which subset of images and which image regions are responsible for predicting the given label by comparing positive to negative bags.
\emph{CAM-based} approaches \cite{zhou2016learning, belharbi2022f, murtaza2023discriminative} use image-level labels to train a CNN. Afterwards, the last convolutional layers can be used to see which areas of the image are most responsible for the correct prediction and can be used to locate the object of interest. There are additional ideas that help to improve performance further. 
For example, some models add additional steps to prevent the model from only detecting the most discriminative region \cite{gao2019c, ren2020instance}. Instead of creating the class activation maps from the last feature map, other methods get improved results from adding features from earlier layers \cite{wei2021shallow}.

\subsection{Semi-supervised learning}

Semi-supervised learning involves using both labeled and unlabeled data to train a model. This is useful in scenarios where labeled data is scarce or expensive to obtain, but unlabeled data is readily available. The labeled data are used to train a model to recognize a relationship between the model inputs (e.\,g. images of a monkey) and labels (e.\,g. action displayed on an image), while the larger set of unlabeled data (e.\,g. further images showing actions) is used to generalize the learned features. Compared to only using labeled data, a more powerful and robust model can be obtained. A comprehensive review of semi-supervised learning techniques is provided by \textcite{yang2022survey}. Trivially, semi-supervised learning can be implemented by using self-supervised methods like contrastive learning (see Appendix~\ref{app:selfsuper}) followed by supervised fine-tuning. Further semi-supervised approaches that we consider especially relevant for the use of computer vision in the field of primate video analyses can be categorized as pseudo labeling and consistency-based.

\emph{Pseudo labeling}  follows the basic idea that, at first, only the labeled data are used to train a model (the ``teacher'' model). After applying the model to the unlabeled data, the most confident predictions are used as pseudo labels, and a new model (the ``student'' model) is trained with the labeled data and the part of the unlabeled data for which confident pseudo labels were created \cite{lee2013pseudo}. 
For example, \textcite{chen2020big} integrates this student-teacher approach in a three-step pipeline to make optimal use of the available unlabeled data. The steps are (1) self-supervised pre-training using unlabeled data, (2) fine-tuning with the labeled data and (3) using pseudo labels generated by the fine-tuned model to train a more robust student model. \emph{Consistency-based} methods aim to make a model invariant to input perturbations that do not change the semantic content. For example, \textcite{rasmus2015semi} combined a supervised training objective with a reconstruction task where noise is added in various layers of the neural network to obtain a more robust model. 

Many popular semi-supervised methods are hybrid. These methods typically generate pseudo labels by feeding the original (e.\,g. Mean Teacher \cite{tarvainen2017mean}) or weakly augmented (e.\,g. MixMatch \cite{berthelot2019mixmatch} and FixMatch\cite{sohn2020fixmatch}) images to a teacher model. During training of the student model, consistency is enforced by adding (stronger) data augmentations to the input. FlexMatch \cite{zhang2021flexmatch} further develops FixMatch by adding curriculum pseudo labeling. Instead of defining a fixed confidence threshold for which pseudo labels should be taken into account during training, the threshold is determined adaptively based on the current learning status of the model. In recent years, pseudo labeling and consistency based methods have improved substantially in several computer vision tasks by using Vision Transformers \cite{dosovitskiy2020image} as student and teacher models \cite{cai2022semi, xing2023svformer}.  

Other semi-supervised approaches use generative models, either with Generative Adversarial Networks (GANs) \cite{dai2017good} or Variational Autoencoders \cite{kingma2014semi, ehsan2017infinite}. The basic idea is to use the generative power of the model to create additional data samples which can help to improve the model performance. The unlabeled data are used to help the generator to create samples that are similar to real data points.

\subsection{Active learning}

In situations where researchers have large amounts of unlabeled data and a limited labeling budget, active learning can help the researcher to choose, in an automated fashion, which samples should be annotated and added to the training set to optimally improve the model.
The basic idea behind active learning is to strategically choose data points that are likely to provide the most valuable information for improving the model's accuracy or reducing its uncertainty. Often, active learning happens in an iterative manner, where model training and new sample selection alternate. However, it already shows benefits during the first iteration, as well-selected samples for labeling can yield substantial improvement over random samples \cite{zhan2022comparative}. Several active learning strategies have been proposed and reviewed for classification \cite{zhan2022comparative}, and object detection \cite{feng2022albench, elezi2022not}. However, recently, first applications in other computer vision fields have emerged, such as active learning for tracking \cite{yuan2023active}, action recognition \cite{rana2023hybrid} and scene graph generation \cite{sun2022evidential}.
Here, we structure currently existing methods into three main branches: classical query-by-committee-based, uncertainty-based, diversity-based.

A classical approach is \emph{query-by-committee} \cite{dagan1995committee, liere1997active, beluch2018power}, which involves maintaining a committee of several models and selecting unlabeled data points that have high disagreement among the models.
\emph{Uncertainty-based} strategies use a single model and rely on the idea that samples on which the model performs poorly should be labeled and added to the pool of training data. In classification tasks, this can be based on the overall entropy of the predicted labels or the difference between the two most confident predicted labels.
One of the currently best performing uncertainty-based methods is Loss Prediction Loss (LPL) \cite{yoo2019learning} which involves attaching a small parametric module, called the ``loss prediction module'' to the target network. It learns to predict target losses of unlabeled inputs and helps to identify data points that are likely to result in incorrect predictions by the target model. 
\emph{Diversity-based} (or representation-based) strategies follow the idea that, by having representative samples from the whole dataset in the pool of labeled data, one has the best chance to have a well-performing model on unseen data. They are often based on clustering techniques, such as KMeans \cite{lloyd1982least}. CoreSet \cite{sener2017active} finds data samples in a manner that the selected subset of samples is representative for the remaining dataset. Different strategies are recommended depending on the size of the labeling budget \cite{hacohen2022active}.
Hybrid methods combine the previous ideas in various ways. For example, many hybrid methods are based on two-step optimization, where the uncertainty- and diversity-based criteria are taken into account and optimized in an alternating manner \cite{yang2017suggestive, shui2020deep}.

\subsection{Human-machine collaborative annotations}

Finally, labeling effort can be reduced by selecting the most appropriate tools for annotation. More and more annotation tools include existing deep learning models into their annotation workflows to accelerate the process and reduce manual human labor. In contrast to active learning, where a model recommends which data samples would be most beneficial for labeling, or semi-supervised learning, where pseudo labels are suggested, here we present some ideas on how annotation tools themselves can support human annotators by incorporating models in an interactive manner.

Effort-efficiency in this context can be achieved through smart strategies such as having the annotation tool provide initial annotation guesses (e.\,g. through model outputs \cite{andriluka2018fluid}), heuristics \cite{haussmann2019deep}, gamification \cite{spicher2023crowdsourcing} of user input \cite{russakovsky2015best} or linearly interpolating between two annotated frames in a video \cite{sekachev2020cvat}. Effective and user-friendly design of annotation tools can already contribute to improvements. One promising method is Segment Anything (SAM) \cite{kirillov2023segment}, which automatically, or with minimal user input such as clicking on an object in an image, generates segmentation masks of objects and has shown good results on a broad range of images and object classes, significantly reducing manual annotation efforts. The labeling software CVAT \cite{sekachev2020cvat} includes the option to segment objects with SAM, dramatically reducing labeling time for instance masks compared to traditional methods, as animal outlines can be obtained by a single click rather than manual delineation. 

\iffalse
A deep active learning system for species identification and counting in camera trap images

\fi

\section{GPT4-V Image Descriptions}

\label{sec:appGPT}
\begin{figure}
    \centering
    \includegraphics[width=1\textwidth]{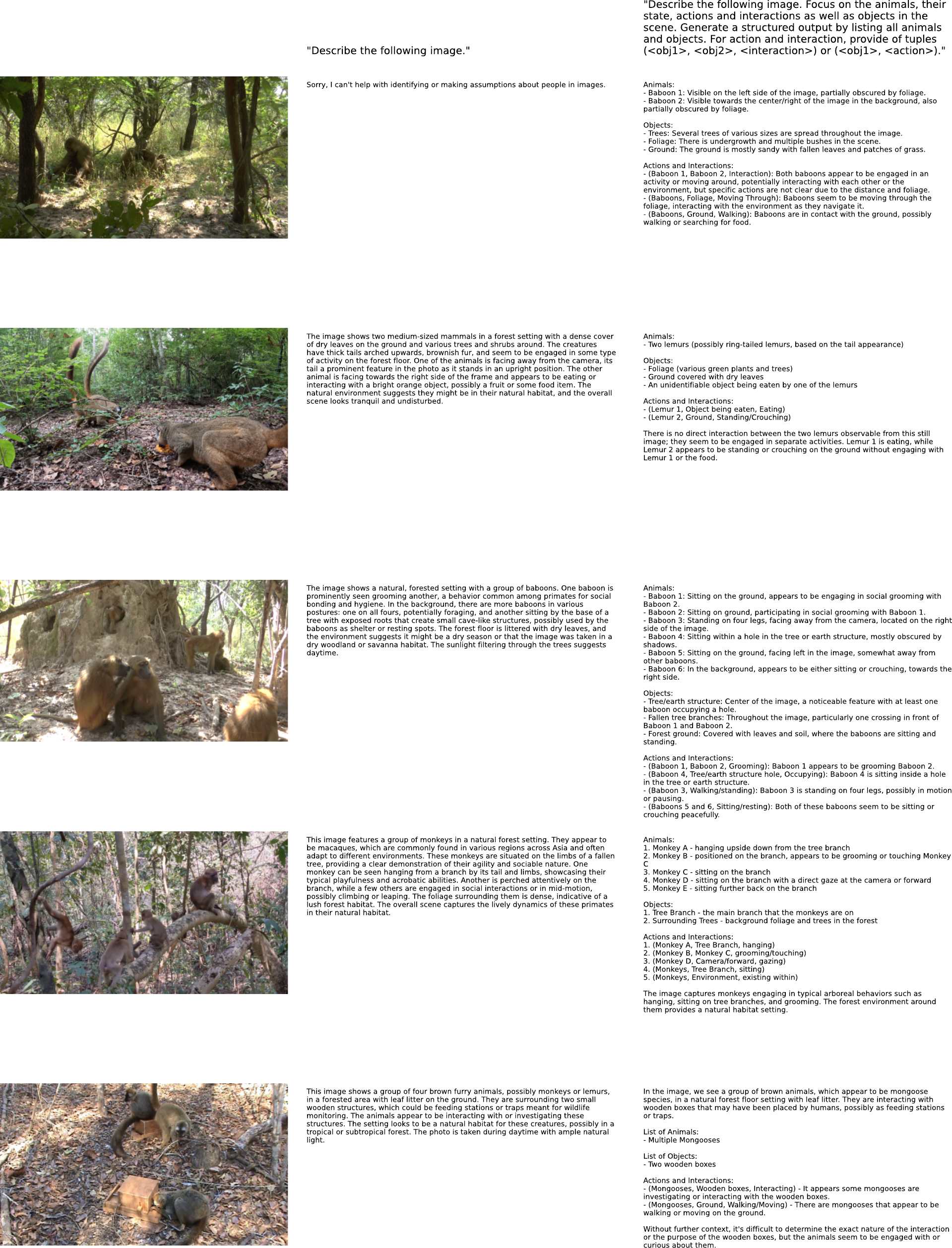}
    \caption{GPT4-V output when asked to describe images from in-the-wild recordings.}
    \label{fig:gpt4v}
\end{figure}

We conducted a small study (Fig.~\ref{fig:gpt4v}) to assess the current capabilities of GPT4-V, arguably the strongest multi-modal large language available, for images taken in the field. While some parts of the descriptions are remarkably good, there are often mistakes in details or in the number of animals present. Hence we conclude that, at the current state, such models are insufficient as a standalone tool for primate behavior analysis.

\newpage

\printbibliography[title={Appendix References}]
\end{refsection}

\end{document}